\documentclass[preprint,12pt,authoryear]{elsarticle}

\makeatletter
\def\ps@pprintTitle{%
 \let\@oddhead\@empty
 \let\@evenhead\@empty
 \let\@oddfoot\@empty
 \let\@evenfoot\@empty
}
\makeatother

\usepackage[a4paper, top=1.2in, bottom=1.2in, left=1in, right=1in, headheight=15pt]{geometry}

\usepackage[utf8]{inputenc}
\usepackage{amssymb}
\usepackage{gensymb}
\usepackage{booktabs}
\usepackage{graphicx}
\usepackage{subcaption}
\usepackage{float}
\usepackage[colorlinks=true, linkcolor=blue, citecolor=blue, urlcolor=blue]{hyperref}
\usepackage[table]{xcolor}
\usepackage{diagbox}
\usepackage{amsmath}
\usepackage{multirow}
\usepackage[round]{natbib}
\usepackage{fancyhdr}
\usepackage{lipsum}  
\usepackage{graphicx}  

\usepackage[symbol]{footmisc}  
\biboptions{round,sort&compress}

\fancypagestyle{plain}{
  \fancyhf{}  
}

\usepackage{microtype}
\usepackage{siunitx}
\usepackage[font=small,labelfont=bf,labelsep=period]{caption}

\begin{document}

\begin{frontmatter}

\title{\textbf{Early-Cycle Current Pulses Enable ML-Based Battery Cycle Life Predictions Across Manufacturers}}

\author[sbaq]{Tyler Sours\footnote[1]{Contributed equally.}}
\author[sbaq]{Shivang Agarwal\footnotemark[1]}
\author[nvnx]{Marc Cormier\footnotemark[1]}
\author[sbaq]{Jordan Crivelli-Decker}
\author[sbaq]{Steffen Ridderbusch}
\author[nvnx]{Stephen L. Glazier}
\author[nvnx]{Connor P. Aiken}
\author[sbaq]{Aayush R. Singh}
\author[sbaq]{Ang Xiao\footnote[2]{Corresponding authors: ang.xiao@sandboxaq.com, omar.allam@sandboxaq.com.}}
\author[sbaq]{Omar Allam\footnotemark[2]}

\address[sbaq]{SandboxAQ, Palo Alto, California 94301, United States}
\address[nvnx]{NOVONIX Battery Technology Solutions, Bedford, Nova Scotia}


\makeatletter
\renewcommand{\abstractname}{}
\makeatother

\begin{abstract}
\noindent Predicting the end-of-life (EOL) of lithium-ion batteries across different manufacturers and operating conditions presents significant challenges due to variations in electrode materials, electrolyte formulations, manufacturing processes, cell formats, and a lack of publicly available data. Methods that construct features solely on voltage-capacity profile data typically fail to generalize across cell chemistries. This study introduces a novel methodology to develop features using Direct Current Internal Resistance (DCIR) measurements from current pulses that span the available state-of-charge range within the first 100 cycles of cell life, enabling more accurate and generalizable EOL predictions.
The use of early-cycle DCIR data captures critical degradation mechanisms and enhances model robustness. Models are shown to successfully predict the number of cycles to EOL, ranging from 700 to 2400 cycles, for unseen cell types containing varied electrode compositions with mean absolute errors (MAE) of 105, 160, and 230 cycles for EOL conditions of 90, 85, and 80 \%, respectively. This generalizability between manufacturers reduces the need for extensive new data collection and retraining, enabling manufacturers to optimize new battery designs using existing datasets and short-term testing. 
Additionally, a novel DCIR-compatible dataset is released as part of ongoing efforts to enrich the growing ecosystem of cycling data and accelerate battery materials development.
\end{abstract}

\end{frontmatter}



\section{Introduction}
\label{introduction}

Lithium-ion batteries are indispensable energy storage devices across a wide range of applications due to their high energy densities, decreasing costs, and extended lifetimes. However, predicting the end of life (EOL) of these batteries based on capacity loss, which is critical to ensuring their reliability and safety, remains a significant challenge due to the complex and nonlinear degradation mechanisms involved. Early and accurate prediction of battery EOL is essential to enable accelerated testing and validation of new battery formulations, allowing manufacturers to optimize chemistries and processes more efficiently. 

\citet{severson2019data} demonstrated that by utilizing early-cycle features from constant-current voltage versus discharge capacity curves, battery life could be predicted with high accuracy within a single cell chemistry. Since then, a wide array of studies have focused on improving EOL predictions \citep{elmahallawy2022comprehensive, ling2022review, paulson2022feature, saxena2022convolutional, ma2020capacity, fei2022early, geslin2023selecting, saxena2021recurrent, chen2022transformer, li2019remaining, rangel2021machine, saxena2015quantifying, zhang2018long, hosen2021battery, attia2021statistical, ng2020predicting}. However, many of these works have relied on features derived from constant-current voltage versus capacity curves, leading to models that perform well only within the specific datasets on which they were trained. When applied to different chemistries or operating conditions, their predictive power diminishes significantly \citep{severson2019data, sulzer2021challenge, finegan2021application, attia2020closed, paulson2024multivariate, finegan2019battery, walker2019decoupling, finegan2019modelling, wang2024open}.
\citet{paulson2022feature} created a machine learning model that generalized across several cathode chemistries, utilizing a dataset drawn from a substantial internal archive of batteries produced in-house. Recently, \citet{rahmanian2024attention} proposed an attention-based recurrent neural network capable of handling diverse chemistries and cycling protocols. While their model demonstrates adaptability, it requires fine-tuning for new chemistries and is reliant on large proprietary datasets. This reliance could potentially limit its practical application in environments with restricted data availability and application to new datasets. For these reasons, there remains a need for battery cycle life prediction models that generalize across manufacturers and operating conditions without relying on data-hungry models.
To add further complexity to the problem, battery performance is influenced not only by electrode chemistry and electrolyte formulation but also by a variety of factors related to cell design and manufacturing processes \citep{baumhofer2014production}. Even cells with nominally similar chemistries can exhibit vastly different behaviors due to differences in manufacturing, including electrode coating thickness, particle size distribution, slurry preparation, and precision in assembly. Such variations can affect the electrochemical performance, aging mechanisms, and overall reliability of a cell, making cross-manufacturer generalization a significantly more complex challenge than cross-chemistry generalization alone. These subtle changes profoundly impact degradation pathways, leading to different rates and patterns of failure. For instance, as a battery cell cycles, degradation processes such as the formation and growth of the solid electrolyte interphase (SEI) on the anode surface, lithium plating, and the accumulation of inactive lithium deposits lead to an increase in interfacial resistance. Concurrently, mechanical degradation of electrode materials, loss of active material, and electrolyte consumption hinder ion and electron mobility, resulting in an overall increase in internal impedance \citep{diao2022degradation, fermin2020identification}.


To address this challenge, the present work employs current pulses across the state-of-charge range, known as Direct Current Internal Resistance (DCIR) measurements, from early cycles to derive key features for predictive modeling. This approach effectively captures underlying degradation processes to enable accurate EOL predictions across cell manufacturers and operating conditions. Using the ensemble of DCIR and traditional voltage-capacity curve features from early cycles, models are developed that maintain high predictive accuracy even with simple linear regressors, achieving intrinsic generalizability without the need for complex algorithms or extensive retraining. Feature selection demonstrates the importance of current pulses in early life to obtain generalizable predictions. Incorporating DCIR measurements into routine cycling protocols offers a straightforward and cost-effective way to enhance model robustness and generalizability across varying manufacturers. 

\section{Cycling Protocol and Dataset Generation}
\label{sec:Data collection and cycling conditions}

\begin{figure}[H]
    \centering
    \includegraphics[width=1.0\textwidth]{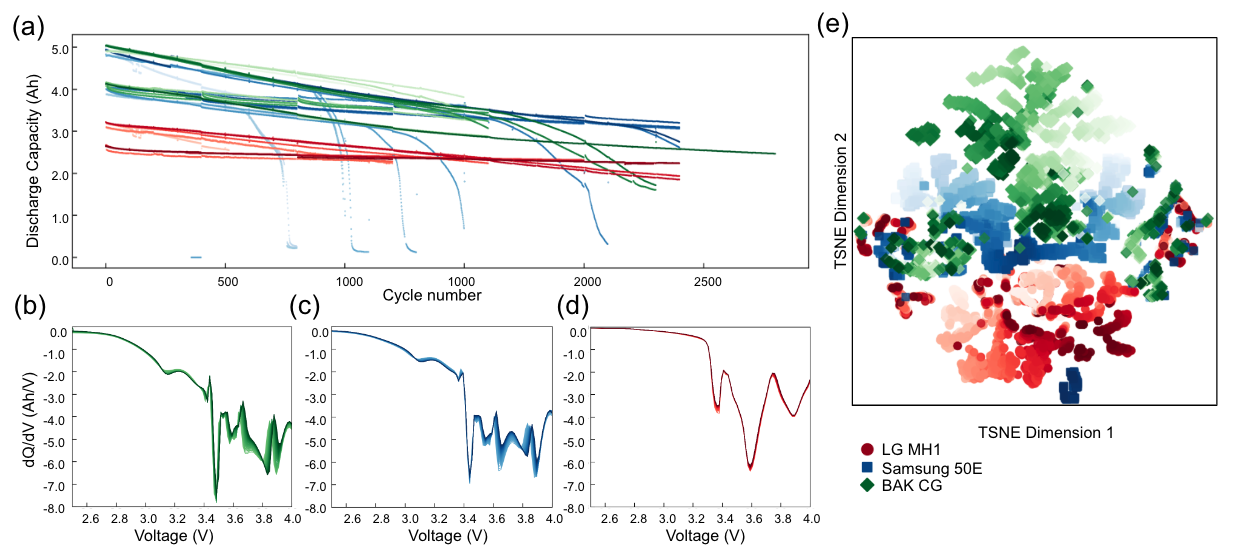}
    \caption{
    \textbf{Distinct cycling behavior across battery manufacturers complicates cross-manufacturer prediction.}
    (a) Discharge capacity as a function of cycle number for battery cells from three manufacturers (LG MH1, Samsung 50E, and BAK CG) under varied operating conditions. Clear differences in degradation rates and capacity loss profiles can be observed across manufacturers.  
    (b-d) Differential capacity (dQ/dV) curves of the first 50 cycles for (b) BAK CG, (c) Samsung 50E, and (d) LG MH1 cells, highlighting the distinct electrochemical signatures of each cell type.  
    (e) t-SNE plot of features derived from cycling data, showing clustering of cells by manufacturer, illustrating the challenge of model generalization across cell types in battery lifetime predictions. 
    }
    \label{fig:Figure1}
\end{figure}

To address the scarcity of standardized datasets across multiple cell chemistries, a comprehensive benchmark dataset acquired from 57 commercial lithium-ion cells from 3 different manufacturers, each with a different cathode-anode combination, is introduced (Fig. \ref{fig:Figure1}). The cells were cycled with a diagnostic protocol that incorporated periodic direct-current internal resistance (DCIR) measurements to track the internal resistance of the battery cells in incremental states-of-charge (SOC) to provide a detailed view of cell degradation and performance, a method widely utilized by OEMs and cell makers.

Battery cycling data were recorded at intervals ranging from a few seconds to a few minutes, depending on the cycling current, generating time-series data for voltage, current, and time. From these fundamental quantities, additional metrics such as capacity and energy were computed. A key metric, the discharge capacity, was tracked across cycles to monitor battery health, where the remaining percentage of initial discharge capacity is referred to as the battery cell's state-of-health (SOH).

The discharge capacity versus cycle number for all cells is illustrated in Fig. \ref{fig:Figure1}a, where cells are colored by manufacturer.  While most cells exhibit linear decay in discharge capacity, some demonstrate non-linear profiles, emphasizing the need for models that can accurately capture these variations. The focus of this work is on predicting the cycle number at which a fixed SOH (e.g., 85\%) is reached, recognizing that different applications may require different SOH thresholds for a cell to be considered at its "end of life". With this in mind, it is noted that while "EOL" may not be the most appropriate terminology for this predictive task, it is used throughout the manuscript for consistency with other works in the field. In this regard, "EOL" refers to a target SOH.

\subsection{Battery cell manufacturers and chemistries}
\label{sec:cells}

The dataset presented in this work is comprised of 57 Li-ion battery cells from 3 different manufacturers, each with a distinct cathode-anode combination. Table \ref{tab:cells} describes the difference between each cell type. The cathode in all cells are considered high-Ni; the Samsung INR21700-50E (Sam-50E) cells contain a Nickel-Cobalt-Aluminum (NCA) cathode \citep{popp2020ante, schmitt2023full} while the BAK N21700CG (BAK-CG) and LG INR18650-MH1 (LG-MH1) cells contain a Nickel-Manganese-Cobalt (NMC) cathode. The anode in all cells is graphite-based; the Sam-50E and BAK-CG cells contain a Graphite-Silicon (Gr-Si) blend anode, while the LG-MH1 cells contain Graphite-only (Gr) anode. 

\begin{table}[ht]
\centering
\begin{tabular}{l l l l c}
    \toprule
    Cell & Cathode & Anode & Format & Nominal capacity \\
    \midrule
    Samsung INR21700-50E & NCA & Gr-Si & 21700 cylinder & 5 Ah \\
    BAK N21700CG & NMC & Gr-Si & 21700 cylinder & 5 Ah \\
    LG INR18650-MH1 & NMC & Gr & 18650 cylinder & 3 Ah \\
    \bottomrule
\end{tabular}
\caption{Manufacturer differences between the three cell types used in this study.}
\label{tab:cells}
\end{table}

These cells differ in their electrode compositions and form factors, which directly influence their degradation behaviors. Fig. \ref{fig:Figure1}a shows that the discharge capacity versus cycle number curves have varying characteristics depending on the manufacturer and cycling conditions; linear, sub-linear, and super-linear decay profiles are observed \citep{attia2022knees}. Fig. \ref{fig:Figure1}b-d demonstrates the differences in electrochemistry between the three cell types, shown by the evolution of the differential capacity profiles over 50 cycles (data collected on a high-resolution NOVONIX UHPC system at 40 °C and a rate of C/10). To visualize the differences in aging characteristics between each cell type, the differential capacities are used for a t-distributed stochastic neighbor embedding (t-SNE), shown in Fig. \ref{fig:Figure1}e. 

By incorporating this diversity, the presented dataset is designed to validate the generalizability of machine learning models across commercial cell chemistries. This approach ensures that the findings presented in this work are not limited to a single cell type but are applicable across a broad spectrum of commercially relevant lithium ion cells.

\subsection{Operating conditions}
\label{sec:doe}

 Table \ref{tab:cell_conditions} summarizes the various cycling conditions that comprise this dataset. Cells were tested within three voltage ranges and at three temperatures to capture a broad spectrum of degradation behaviors. The 3.3 - 4.2 V range was chosen to avoid accessing capacity from the Si component of the Gr-Si negative electrodes, which occurs at lower voltages where Li alloys with Si. For this reason, the Sam-50E and BAK-CG cells were cycled within a 3.3 V - 4.2 V range but the LG-MH1 cells were not because they do not contain any Si. Cycling at 60 °C was conducted exclusively within the 2.5 V - 4.2 V range.

Each set of operational parameters (temperature, voltage range, and manufacturer) included three identical cells to account for cell-to-cell variability and the influence of different cycling conditions. This comprehensive approach supports the development of predictive models that are robust across a wide range of real-world scenarios.

\begin{table}[ht]
\centering
\resizebox{0.90\textwidth}{!}{
\begin{tabular}{c c c c}
    \toprule
    \multirow{2}{*}{T (\si{\degreeCelsius})} & \multicolumn{3}{c}{Voltage Range (V)} \\
    \cmidrule(lr){2-4}
     & 2.5--4.06 & 2.5--4.2 & 3.3--4.2 \\
    \midrule
    23 & Sam-50E, BAK-CG, LG-MH1 & Sam-50E, BAK-CG, LG-MH1 & Sam-50E, BAK-CG \\
    40 & Sam-50E, BAK-CG, LG-MH1 & Sam-50E, BAK-CG, LG-MH1 & Sam-50E, BAK-CG \\
    60 & -- & Sam-50E, LG-MH1 & -- \\
    \bottomrule
\end{tabular}
}
\caption{Cell testing conditions used in this study, grouped by temperature and voltage range.}
\label{tab:cell_conditions}
\end{table}

All cells were cycled at a constant-current rate of C/3, with a constant-voltage hold at the top-of-charge. To monitor and assess cell health, a diagnostic sequence was employed every 100 cycles, starting at the 2nd cycle, that included two low-rate cycles at C/10 followed by a 10-step DCIR measurement. The DCIR measurement was performed using 10 current pulses at equally-spaced SOCs, decrementing in 10\% intervals from 100\% to 10\%. Each pulse had a 10-second duration at a current equivalent to a 1C full discharge and was followed by a 15 minute relaxation period, a design that minimizes disruption to the cell's normal cycling and ensures accurate resistance measurements without compromising cell performance.

This protocol efficiently captures dynamic performance and overall battery health under realistic operating conditions, providing a rich dataset for developing predictive models that can generalize across various chemistries and conditions.

\section{Machine Learning Approach}
\label{Machine Learning Approach}

    \begin{figure}[h]
    	\centering 
    	\includegraphics[width=1.0\textwidth]{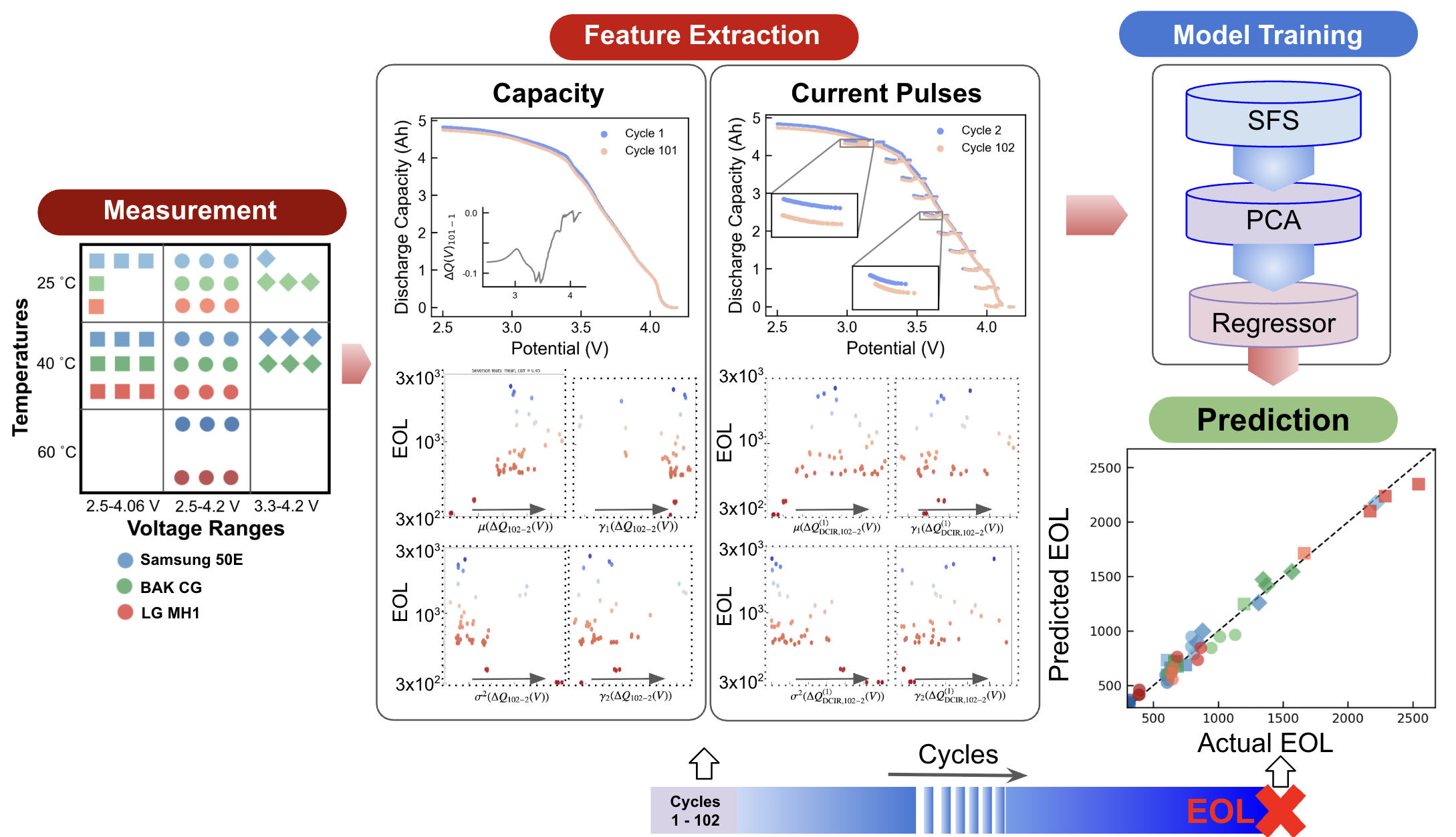}	
        \caption{
        \textbf{Schematic representation of feature extraction for battery cycle life prediction.} Starting with 102 cycles of measured time-series data for different cell types and operating conditions, features are computed from differences in voltage-capacity profiles using both the full voltage range and current pulses, and effective resistances (DCIR). The feature set dimensions are reduced using sequential feature selection (SFS) and principle component analysis (PCA). Models are trained to predict the number of cycles to a target end-of-life EOL condition.
        }
    	\label{fig:Figure2}
    \end{figure}

\subsection{Feature set}
\label{feature_set}

Comprehensive feature sets were extracted from time-series data of the first 102 cycles (Fig. \ref{fig:Figure2}). The feature sets were down-selected before assessing model performance, and section \ref{subsec:lomo} presents associated implications, particularly how selected features preferentially correspond to electrochemically meaningful quantities. The range of features considered in this work are:

\begin{enumerate}
    \item \textbf{$\Delta Q(V)$ Features} --- Following the work of \citet{severson2019data}, univariate time-series features were extracted from the difference between discharge capacity vs. potential curves, $Q(V)$, for two specific cycles (in this case, cycles 101 and 1, corresponding to the slower C/10 rate step prior to the DCIR checkup cycle). The extracted features include skewness ($\gamma_1$), kurtosis ($\gamma_2$), variance ($\sigma^2$), mean ($\mu$), minimum ($\min$), and maximum ($\max$) of the function:
    \[
    \Delta Q_{101-1}(V) = Q_{101}(V) - Q_{1}(V)
    \]
    where $Q_{n}(V)$ represents the $Q(V)$ curve at cycle $n$. These statistical features capture the distributional and shape changes in $Q(V)$ curves. Fig. \ref{fig:Figure2} depicts sample $Q_{1}(V)$ and $Q_{101}(V)$ curves, with the inset showing the corresponding $\Delta Q_{101-1}(V)$.
    
    \item \textbf{$R_{\text{DCIR}}$ Features} --- Each DCIR measurement involves applying a current pulse at a specific state of charge (SOC) during a discharge step. The response to a current pulse is a voltage drop that allows calculating an effective internal resistance: $R = \Delta V / I$. Included in the feature sets were resistance values extracted for each of the 10 current pulses during cycles 2 and 102. Fig. \ref{fig:Figure2} shows a sample DCIR pulse sequence. 

    \item \textbf{$\Delta Q_{\text{DCIR}}(V)$ Features} --- Recognizing that each current pulse from the DCIR measurements from partial $Q(V)$ curves, the methodology from \citet{severson2019data} was extended to calculate the same set of univariate time-series statistics (mean, minimum, maximum, variance, skewness, and kurtosis) on these partial curves. Considering partial $Q(V)$ curves across the 10 SOC levels, between cycles 2 and 102, generated 60 additional features, which are referred to as $\Delta Q^{(k)}_{\text{DCIR}}(V)$ features, where $k$ represents the pulse number ($k=0$ is the pulse at 100 \% SOC). This approach enriches the feature set by capturing nuances in voltage response at various SOC levels. 

    \item \textbf{Linear Fit Features} --- To capture the overall trend in discharge capacity over time, a linear fit is applied to the discharge capacity vs. cycle number curve between two early cycles (e.g., cycles 30 and 99). The slope ($p_1$) and intercept ($p_2$) of this linear fit were used as features. These features provide insights into the rate of capacity fade and the initial state of the battery. Although more complex fits were explored, they did not significantly improve model performance and were therefore not included in the final feature set.

\end{enumerate}

These features were combined to train various machine learning models. For regression-based modeling, the best performance was achieved using a simple linear Elastic Net model. While more complex models were tested (as detailed in the Appendix), the Elastic Net model was selected due to its regularization properties, which help mitigate overfitting, especially when using a relatively small dataset or when the features exhibit multi-collinearity. This is particularly important given the nature of our feature sets, which include correlated variables from both traditional voltage-capacity profile features and those derived from DCIR measurements. Importantly, as a pre-processing step, Sequential Feature Selection (SFS) was applied to the high dimensional feature set to refine the input features and reduce the dimensionality of the feature space. The resulting feature space dimension was dependent on the cross-validation scheme (discussed in the following paragraphs). Principal Component Analysis (PCA) was then applied to the dimensionally-reduced feature sets. This method demonstrates the robustness of the selected features which achieve high prediction accuracy without the need for complex or resource-intensive models. Although the models trained in this work show strong predictive accuracy, it is noted that a larger and more diverse dataset is essential to further validate these findings and ensure broader applicability. While cross-validation provides a robust estimate of model performance, the addition of an independent holdout dataset would offer a more definitive evaluation of the model's capacity to generalize across different chemistries, manufacturers, and operating conditions. Due to the limited dataset size, particularly considering the use of triplicate cells, a holdout set was not deemed feasible for this work. Instead, various cross-validation strategies were considered to assess the generalizability of the feature set.

\subsection{Cross-validation strategies}
\label{sec:cv_schemes}

To evaluate the generalization performance of the trained models, three distinct cross-validation strategies were employed, each focusing on different aspects of the dataset variability.

First, since the dataset is composed of cells grouped into triplicates, where each triplicate is a set of three cells cycled under identical conditions (i.e., a specific combination of manufacturer, temperature, and voltage range), a \textit{leave one triplicate out} (LOTO) cross-validation strategy was adopted to assess the model's ability to generalize across unique samples.  In this approach, the model was trained on all but one triplicate, and the remaining triplicates were used for testing. This process was repeated until all triplicates were used as test sets, thus evaluating the model across a diverse range of operating conditions and cell types. By excluding entire triplicates during each fold, this method tests the model's ability to predict cycle life for a particular cell type and unseen operating condition. This strategy maximizes the utility of the dataset and provides an assessment of the model’s generalization capabilities across temperature, voltage, and manufacturer combinations.

Second, the LOTO strategy was extended to \emph{leave one operating condition out} (LOOCO) in which folds are created by omitting from model training all cells tested under a particular combination of temperature and voltage range. This approach assesses how well the feature set generalizes to unseen operating conditions. 

Third, \emph{leave one manufacturer out} (LOMO) cross-validation was performed to assess the model's ability to generalize across different cell types. This was achieved by training the model on cells from two manufacturers and testing it on cells from the third. It is important to emphasize that with this approach, the models were tested on unseen cell types across the various temperature and voltage range combinations. This approach evaluates the capacity of the feature set to generalize to new cell types. It is important to note that the cell types in this dataset have in common Ni-based cathodes and Gr-based anodes, thus the LOMO strategy does not validate across other chemistry families, such as LFP, for example. However, it is worth re-iterating that the three cell types considered here were each distinct in their exact cathode-anode compositions (e.g, NCA and NMC, or Gr-Si and Gr). The LOMO strategy is an important consideration for assessing model performance because manufacturing processes introduce variability in electrode design, assembly, and material quality, thus ensuring robustness across different production sources.

For completeness, a standard randomized 5-fold cross-validation was also performed, and the results are presented in the Appendix. In general, predictions are better using 5-fold cross-validation because the models were trained on 1 or more from a triplicate set, while predicting on a cell from the same set. The results presented in this manuscript focus on the cross-validation strategies discussed above which are tailored to the specific challenges of generalization across operating conditions and manufacturers. 

\section{Results and Discussion}
\label{sec:results_discussion}

To establish a performance baseline for model performance across operating conditions and cell types, predictions are compared against features computed from the differences in voltage vs capacity profiles between cycles 101 and 1 ($\Delta Q_{101-1}(V)$, described in section \ref{feature_set}) and using the \emph{leave one triplicate out} cross-validation strategy (LOTO, described in section \ref{sec:cv_schemes}). These features have been well-studied in the literature \cite{severson2019data} thus provide a good reference point. In the following discussion, comparisons are made across feature sets and cross-validation strategies to assess model generalizabilty.

\begin{figure}[H]
    \centering
    \includegraphics[width=1.0\textwidth]{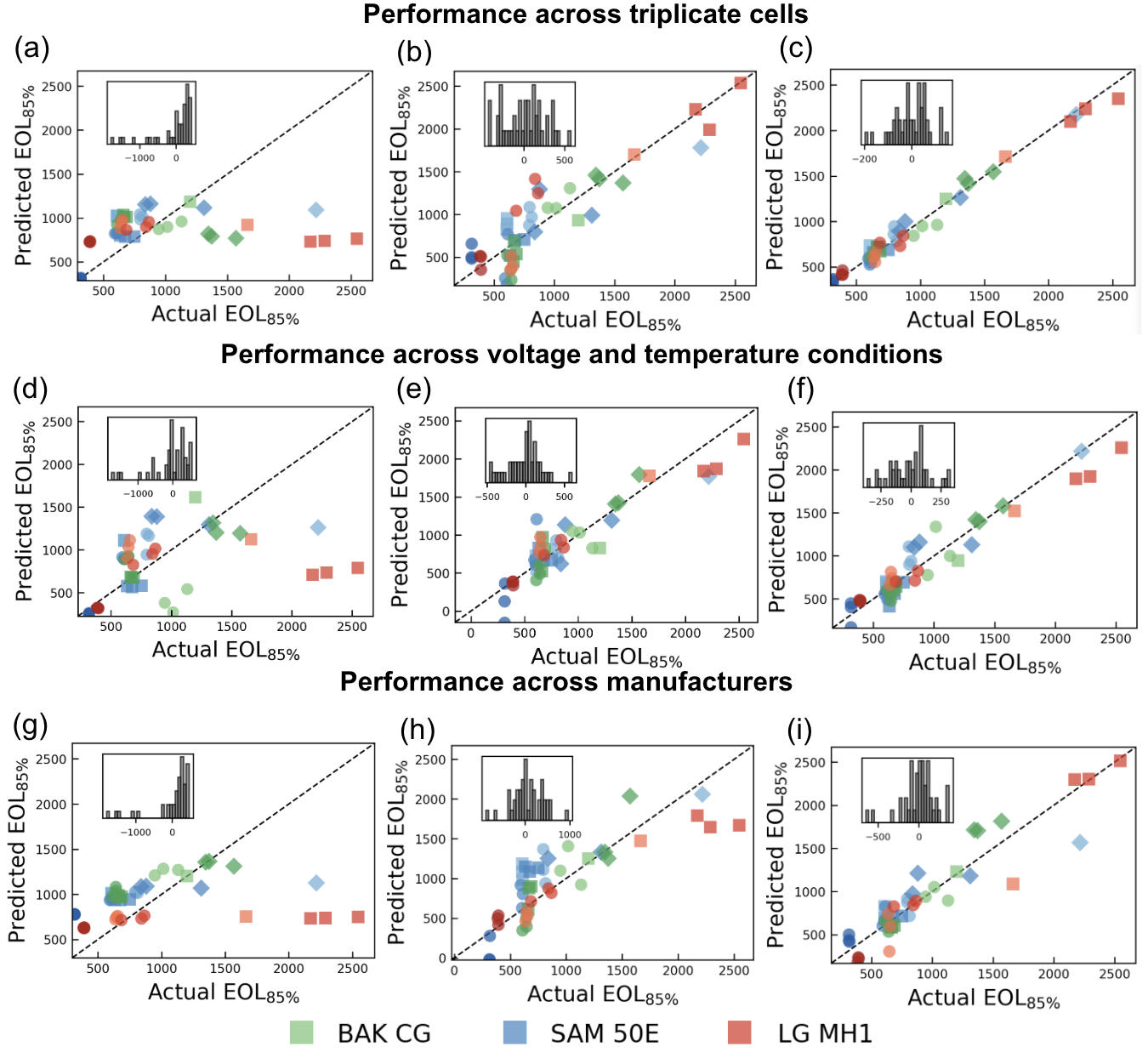}
    \caption{
    \textbf{Predicted vs. actual cycles to EOL for different feature sets from an elastic net regressor across varying conditions.}
    Panels (a) (d) and (g) depict predictions using the feature set from \citet{severson2019data}. Panels (b), (e) and (h) show predictions using a downselected set of features from DCIR current pulse differences at various states of charge between cycles 102 and 2. Panels (c), (f), and (i) show results for a down-selection of all features considered in this work.
    The top, middle, and bottom rows correspond to \emph{leave one triplicate out}, \emph{leave one operating condition out}, and \emph{leave one manufacturer out} cross validation strategies, respectively.
    }
    \label{fig:Figure3_4}
\end{figure}

\begin{figure}[H]
    \centering
    \includegraphics[width=0.45\textwidth]{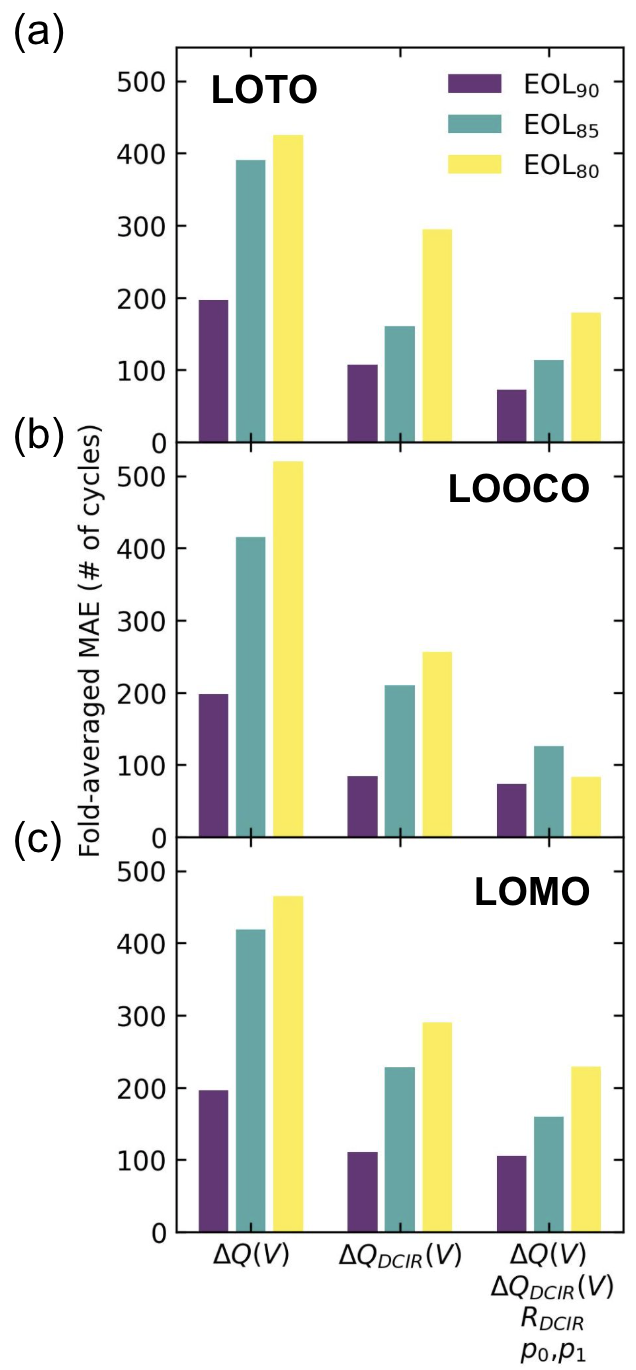}
    \caption{
    \textbf{Prediction Mean Absolute Error across feature sets, end-of-life thresholds, and cross-validation strategies.}
    Mean Absolute Error (MAE) against feature set averaged over all cross-validation folds using (a) \emph{leave-one triplicate out} (LOTO), (b) \emph{leave-one operating condition out} (LOOCO), and (c) \emph{leave-one manufacturer out} (LOMO) cross-validation strategies and for 90\% (purple), 85\% (green), and 80\% (yellow) end-of-life thresholds.   
    }
    \label{fig:features-cv}
\end{figure}

Fig. \ref{fig:Figure3_4}a-c shows parity plots comparing predicted and actual EOL$_{85}$ with the LOTO cross-validation strategy. Fig. \ref{fig:Figure3_4}a,b,c respectively show predictions made with models trained on features computed from voltage profile differences ($\Delta Q_{101-1}(V)$), current pulse differences ($\Delta Q_{DCIR} (V)$), and a combination of all features that additionally consider direct-current internal resistance ($R_{DCIR}$) and linear fit parameters to the lifetime curves. In each case, the feature sets are down-selected to 20 features. The predictions systematically improve when current pulses are included in the feature selection. 
Fig. \ref{fig:features-cv}a shows the MAE averaged over all cross-validation folds for the LOTO cross-validation strategy and for three EOL thresholds (90\%, 85\%, and 80\%). The green squares (EOL$_{85}$) correspond to the average MAE for the parity plots of fig. \ref{fig:Figure3_4}a-c. The average MAE is reduced from more than 400 cycles to less than 200 cycles for the EOL$_{80}$ predictions and from ~400 cycles to nearly 100 cycles for the EOL$_{85}$ predictions. The differences in model MAE for the various EOL conditions can be attributed to two factors: i) some cells did not reach the EOL$_{80}$ or EOL$_{85}$ threshold which means there were fewer cells in then training set, and ii) in earlier cycles the lifetime curves are highly linear, whereas in later life some of these curves exhibit non-linear behavior that makes predictions more challenging. Nevertheless, the inclusion of features from current pulses ($\Delta Q_{DCIR}(V)$ and $R_{DCIR}$) reduces the prediction error by a factor of 2 to 3. 

\subsection{Cross voltage and temperature performance}
\label{subsec:loto}

Fig. \ref{fig:Figure3_4}d-f depicts how well model predictions generalize to unseen temperatures and voltage ranges (LOOCO cross-validation described in section \ref{sec:cv_schemes}) across feature sets and using an EOL threshold of 85\%.  The same three sets of derived features as fig. \ref{fig:Figure3_4}a-c were used but down-selected to 12 features since each cross-validation fold contained fewer samples. Predictions improve significantly with the inclusion of features computed from current pulses; fig. \ref{fig:Figure3_4}e shows a noticeable tightening of the predicted vs actual EOL data to the ideal line (black-dotted) compared to fig. \ref{fig:Figure3_4}d. This particular comparison is interesting because in both scenarios univariate time-series features from differences in voltage vs capacity profiles are used, thus highlighting that current pulses spanning the available SOC range encode important degradation information in the electrochemical signature. To a greater degree, injecting phenomenological insight in the feature generation (fig. \ref{fig:Figure3_4}f) allows for simple feature construction from current pulses by computing an effective resistance ($R_{DCIR}$) for each pulse and improves the prediction error further. Section \ref{sec:indicators} elaborates on the importance of various features. Prediction errors are summarized in fig. \ref{fig:features-cv}b, showing the averaged MAE for the LOOCO strategy and the three EOL conditions. The green squares correspond to panels d-f in fig. \ref{fig:Figure3_4}. When attempting to predict EOL with an 85\% capacity threshold (green squares), the standard $\Delta Q(V)$ features yield an average MAE of over 400 cycles, while the optimized feature set that leverages current pulses yields an MAE just over 100 cycles; nearly four-fold reduction in prediction error. This demonstrates that features based on current pulses capture voltage- and temperature-dependent degradation modes that are not discernible from single current voltage profiles.

\subsection{Cross cell manufacturer performance}
\label{subsec:lomo}

Fig. \ref{fig:Figure3_4}g,h,i demonstrate how features based on current pulses enable models to generalize well to new cell types by using the \emph{leave one manufacturer out} (LOMO) cross-validation strategy presented in section \ref{sec:cv_schemes}. In each of the three panels, the same feature sets as fig. \ref{fig:Figure3_4}a,b,c were used but down-selected to 12 features because each cross-validation fold contained fewer samples compared to the LOTO strategy. Similarly to generalization to new operating conditions, the inclusion of univariate time-series features based on current pulses (fig. \ref{fig:Figure3_4}h) improves predictions drastically. The inclusion of features based on effective resistances ($R_{DCIR}$) further tightens the cycle life predictions to the actual values (fig. \ref{fig:Figure3_4}i). Fig. \ref{fig:features-cv}c summarizes predictions using the LOMO cross-validation strategy for the three EOL thresholds and three feature sets considered. The EOL$_{85}$ data (green squares) correspond to the average MAE from fig. \ref{fig:Figure3_4}g,h,i. For cell cycle life degradation beyond 90\%, the improvement in prediction error when current pulse-based features are included is accentuated; the averaged MAE for both EOL$_{85}$ and EOL$_{80}$ are reduced by more than half. Considering that these cycle life prediction errors correspond to cells types that were unseen in training (within each respective cross-validation fold), the inclusion of current pulses, and to a greater extent effective resistances, leads to models that encode information on degradation processes not unique to a single cell chemistry.

\subsection{Model generalizability summary}
\label{subsec:mod_gen}

The averaged MAE summarized in fig. \ref{fig:features-cv} illustrates the importance of encoding the electrochemical information that spans the entire SOC range, with the inclusion of pulse voltage-capacity profiles, and further illustrates the importance of injecting phenomenology in feature generation, with the inclusion of effective resistances. Two points of interests are discussed with regards to the current work: I) the increase in feature set dimension with the inclusion of current pulses, and II) the prediction errors of the voltage-capacity profile features ($\Delta Q(V)$). I) after down-selection, the feature space increases dimension from 6 to 20 (LOTO), from 6 to 12 (LOOCO), and from 6 to 10 (LOMO), thus prediction errors are expected to decrease. However, this is counter-balanced by II) examining fig. \ref{fig:Figure3_4}a,d,g again, it is observed that $\Delta Q(V)$ features make predictions within a band of values in which short-lived cells are over-predicted and long-lived cells are under-predicted, not effectively capturing trends. It can thus be postulated that by including additional long-lived cells the MAE would get inflated. This is not a phenomena solely attributable to feature space dimension, rather it shows the importance of resolving features across the entire available SOC range.

Of particular importance is how close the the prediction errors are in fig. \ref{fig:features-cv}a,b,c for each EOL thresholds and all three cross-validation strategies when considering all four feature sets ($\Delta Q(V)$, $\Delta Q_{DCIR}(V)$, $R_{DCIR}$, $p_0$, $p_1$). It is recalled that since some cells degraded slowly, the number of samples available for training decreased as the EOL threshold decreased; fewer cells reached 80\% SOH compared to 85\%, and again compared to 90\%. This could certainly contribute to the general increase in prediction error as the EOL threshold is reduced. Nevertheless, considering the varied degradation trends exhibited within this dataset (fig. \ref{fig:Figure1}), the accurate and consistent cycle life predictions made by models that leverage features from current pulses and phenomenological insight across each EOL condition suggests that different degradation modes are encoded by these features, allowing for accurate predictions even during rapid cell failure. For example, sudden cell failure is generally accompanied with a rapid increase in internal resistance, which makes evident why cycle life prediction models that use $\Delta Q(V)$ features fail to generalize across degradation modes: the detailed electrochemical signature is overlooked. Developing features from current pulses that span the available SOC range appears to capture detailed differences in degradation such as solid-electrolyte interphase (SEI) growth, active material loss due to particle cracking or structural transformations, and electrolyte salt and solvent depletion. For example, even within a single cell type, differences in cycling voltage can have a dramatic impact on degradation; a Si-containing anode (Sam50E, BakCG) can operate effectively as a graphite-only anode if the lower voltage limit is sufficiently high, mitigating Si-related degradation modes. Thus, varying operating conditions, cell chemistries, and manufacturing processes lead to differences in degradation modes, and the ability to predict cycle life across these factors provides evidence of the rich information encoded in features computed from current pulses that span the entire SOC range.

\section{Electrochemical Indicators and Their Role in Generalization}
\label{sec:indicators}

Differences in battery cell manufacturing such as electrode composition, electrode slurry recipes, electrode coating thickness and uniformity, cell form factors, and assembly precision, introduce significant complexity for cycle life prediction. Subtle differences between cells can manifest during cycling and aging as changes in internal resistance, capacity fade, and overall cell degradation. Internal resistance, in particular, can serve as an indicator of underlying degradation mechanisms, including solid electrolyte interphase (SEI) growth, electrode particle fracture, and lithium plating. These degradation processes influence the electrochemical behavior of the cell, which in turn impacts its performance and lifetime. The following discussion discerns feature importance and offers connection to electrochemical phenomena.

\begin{figure}[h]
    \centering
    \includegraphics[width=1.0\textwidth]{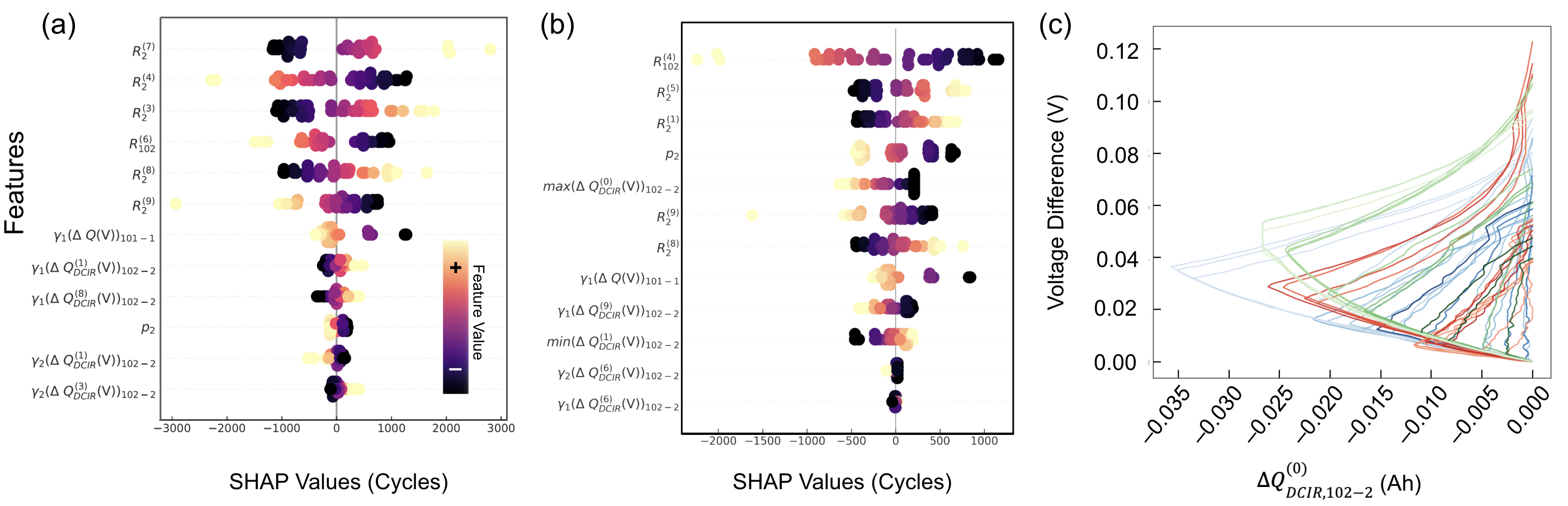}
    \caption{
    \textbf{Feature-based drivers of model performance.}
    SHAP values (in cycles) showing the contribution of various features to the prediction of EOL$_{85}$ for (a) \emph{leave one operating condition out} and (b) \emph{leave one manufacturer out} cross-validation strategies. The features are ordered from top to bottom by relative importance for model prediction. Positive SHAP values indicate a positive contribution to longer predicted cycle life, while negative values imply a reduction in predicted cycle life. The color gradient corresponds to the magnitude of feature values, where darker colors indicate higher values.  
    (c) Voltage difference plotted against $\Delta Q_{DCIR,102-2}^{(0)}$ for cells from three different manufacturers (LG - red, Samsung - blue, BAK - green). Darker colors correspond to a greater number of cycles to EOL.
    }
    \label{fig:Figure5}
\end{figure}

Fig. \ref{fig:Figure5} shows that resistance-based features ($R^{(i)}_{102}$, $R^{(j)}_{2}$) play a critical role in predicting cycle life, contributing substantially to model performance under various operating conditions (Fig. \ref{fig:Figure5}a) and across different manufacturers (Fig. \ref{fig:Figure5}b). These features track changes in SOC-dependent cell resistance due to degradation mechanisms such as solid electrolyte interphase (SEI) growth, electrolyte oxidation, lithium plating, cathode particle surface reconstruction, and poor electrode connectivity that can manifest across different chemistries and manufacturing processes. It is important to realize that the individual SHAP values in Fig. \ref{fig:Figure5}a-b should not be directly extended to electrochemical analysis; models encode relationships between the 12 values to make predictions. Nevertheless, the relative importance of features does indicate that resistance-based features derived from current pulses provide rich information with generalizable predictive power. Fig. \ref{fig:Figure5}c illustrates that cells with a greater number of cycles at EOL$_{85}$, indicated by the darker colors, exhibit smaller $\Delta Q_{DCIR,102-2}^{(0)}$. This suggests that features based on current pulses across the SOC range can effectively differentiate degradation behaviors even when the underlying cell chemistries and manufacturing processes vary significantly. 

In contrast, features based on differences in voltage-capacity profiles primarily track shifts in capacity fade but fail to capture early-stage degradation mechanisms that lead to internal resistance growth, which can occur well before significant capacity loss is observed. This limitation becomes particularly evident when attempting to generalize models across different manufacturers, as subtle variations in manufacturing processes (e.g., electrode coating uniformity or electrolyte formulation) can result in different onset points and rates of degradation. The ability of resistance-based features that span the available SOC range to generalize across varied conditions highlights their importance in achieving reliable lifetime predictions.

Overall, the ability of DCIR features ($R^{(i)}_{DCIR}$) to reflect internal resistance growth, contributes significantly to the model’s robustness in generalizing across cell types and operating conditions. These features effectively capture early-onset degradation processes, ensuring more reliable battery lifetime predictions despite varied operating conditions, cell chemistries, and manufacturing processes.

\section{Conclusion}
\label{conclusion}

This work demonstrates the efficacy of incorporating features from current pulse measurements into data-driven models for the prediction of Li-ion battery cycle life, addressing limitations associated with traditional voltage-capacity-based features. While these conventional features capture broad degradation trends, they fail to generalize effectively across different temperatures, voltage ranges, and cell manufacturers due to variations in cell design, electrode composition, and manufacturing processes. The inclusion of DCIR data, particularly in early cycles, allows for the detection of internal resistance growth, which is closely tied to key degradation mechanisms. Results presented in this work show that this integrated approach significantly improves prediction accuracy and generalizes well across different manufacturers and chemistries. Moreover, this methodology has the potential to allow manufacturers to adapt their battery compositions and leverage pre-existing data for predictive modeling without the need to build entirely new datasets, facilitating rapid innovation and optimization of new cell designs. Incorporating DCIR-based features into predictive models could thus provide a more robust and efficient framework for EOL estimation that can accelerate research and development and potentially provide early warnings for in-field applications.

\section{Methods}
\subsection{Data generation}
57 ready-for-use commercial Li-ions cells were obtained for this work. Included were: 21 Samsung INR21700-50E, 21 BAK N21700CG, and 15 LG INR18650-MH1 cells. Cell operating conditions included 3 voltage ranges (2.5 - 4.06 V, 2.5 - 4.2 V, and 3.3 - 4.2 V) and three temperatures (23, 40, and 60 °C). Tests at 60 °C were only performed within the 2.5 - 4.2 V range and the LG INR18650-MH1 cells were not tested within the 3.3 - 4.2 V range. Tests were carried out with triplicate cells (3 cells from the same manufacturer operated at the same conditions and with the same charge-discharge protocol). The experiment design thus consisted of 19 unique samples (combination of cell type, temperature, and voltage range). Five cells experienced early failure with internal current interruption and were omitted from the dataset and anlysis: 1 Samsung INR21700-50E (23 °C, 3.3 - 4.2 V), 1 BAK N21700CG (23 °C, 2.5 - 4.06 V), and 3 BAK N217000CG (60 °C, 2.5 - 4.2 V). Additionally, two LG INR18650-MH1 (25 °C, 2.5 - 4.06 V) cells never degraded to 90 \% or lower and were ommitted from the dataset. Cell cycling was carried out on NEWARE BTS-4000 cyclers. For cells tested at 40 °C, the temperature was maintained using in-house customized industrial thermal chambers, with estimated maximum temperature fluctuations of 1 °C. For cells tested at 60 °C, the temperature was maintained with NOVONIX thermal chambers, with estimated maximum temperature fluctuations of less than 0.1 °C. Cells tested at 23 °C were in a temperature-controlled warehouse with estimated maximum temperature fluctuations of 1 °C. The charge-discharge sequence employed for all cells follows a 100 cycle pattern: C/10 charge, C/10 discharge, C/10 charge, C/10 discharge with 1C-equivalent pulses followed by a 15 minute rest (open-circuit) each 10 \% SOC, 98 cycles at C/3 (constant-current charge with a constant-voltage hold at top-of-charge terminating when the current is less than C/20 rate equivalent, followed by constant current discharge). Sample protocol files have been included in the data repository listed in the Data Availability section.

\subsection{Data processing}
The files output by the cycler systems were processed and stored in the Apache Arrow Feather file format. Some quantities were computed during processing but the original data was preserved through the file store conversion. In the data repository quoted in the Data Availability section, only the Feather files are made available because they contain the original data in addition to all computed quantities. Aggregated quantities (per-cycle metrics) are stored in a separate Feather file. Data dictionaries are provided in the open-access data repository.  

\subsection{Feature generation}
The present work focuses on systematically deriving features from the first 102 cycles, encompassing both full discharge curves and discrete voltage responses during DCIR current pulses. Following the methodology of ~\cite{severson2019data}, a structured interpolation scheme was employed for the capacity--voltage data to facilitate cycle-to-cycle comparisons and ultimately extract statistical descriptors that capture changes in the battery’s state of health.

\noindent \textbf{Traditional Capacity-based features}. For each cycle, the discharge capacity ($Q$) was measured as a function of voltage ($V$). To enable a consistent pointwise comparison between cycles, a voltage window was defined spanning the full discharge range for each cell (e.g., 2.5--4.2\,V), over which linear interpolation of $Q(V)$ was performed. Each $Q(V)$ curve was sampled at a uniform voltage grid, ensuring that capacity data from different cycles could be directly subtracted at the same voltage points:
\[
Q_{n}(V) \;\longrightarrow\; \widetilde{Q}_{n}(v_i),
\]
where $\widetilde{Q}_{n}(v_i)$ denotes the interpolated capacity at the $i$-th voltage point $v_i$.

The difference between two representative cycles $m$ and $n$ was then computed:
\[
\Delta Q_{m-n}(v_i) \;=\; \widetilde{Q}_{m}(v_i)\;-\;\widetilde{Q}_{n}(v_i).
\]
In this study, $\Delta Q_{101-1}(V)$ was selected to capture early-to-later cycle changes in capacity at the slower C/10 discharge rate. Following ~\cite{severson2019data}, a suite of univariate statistical descriptors was extracted from $\Delta Q(V)$, including the mean ($\mu$), minimum ($\min$), maximum ($\max$), variance ($\sigma^2$), skewness ($\gamma_1$), and kurtosis ($\gamma_2$). These descriptors summarize the shape and distributional changes in the discharge curves over time, reflecting diverse degradation mechanisms (e.g., loss of active material, shifts in electrode kinetics).

\noindent \textbf{DCIR-based features.} During the checkup cycles, DCIR measurements were performed by applying short current pulses at specific state-of-charge (SOC) levels and recording the resultant voltage drop. The effective internal resistance, $R_{\text{DCIR}} = \Delta V / \Delta I$, was recorded for each of the 10 SOC checkpoints in both an early cycle (cycle~2) and a later cycle (cycle~102). These measured resistances were used directly as features ($R_{\text{DCIR}, k}$, where $k \in \{1,2,\ldots,10\}$).

 In addition to the resistance measurements, the current pulses in each DCIR checkup define partial discharge segments, each providing a localized $Q(V)$ relationship in a narrower SOC range. By concatenating the voltage and capacity data within each of these partial segments and using the same interpolation approach described above, partial $Q(V)$ curves were formed for each pulse number $k$. The difference between the interpolated partial capacity curves in cycles~102 and~2, $\Delta Q_{\text{DCIR},102-2}^{(k)}(V)$, was then computed for each SOC checkpoint. The same univariate statistical descriptors (mean, minimum, maximum, variance, skewness, kurtosis) were extracted for each $\Delta Q_{\text{DCIR}}^{(k)}(V)$. This yielded 60 additional features (6 statistics~$\times$~10 pulses), collectively referred to as $\Delta Q_{\text{DCIR}}^{(k)}$ features. These metrics capture localized capacity evolution at various SOC levels, enhancing sensitivity to distinct degradation processes that may manifest in specific regions of the voltage curve.

\noindent \textbf{Lifetime Curve Features.}
To capture broad trends in capacity fade, a linear fit of the discharge capacity vs.\ cycle number was applied from an early to an intermediate stage of cycling (e.g., cycles~30--99). Let $Q_{\mathrm{dch}}(c)$ denote the capacity at cycle $c$. The linear model was:
\[
Q_{\mathrm{dch}}(c) \;=\; p_1\,c + p_2,
\]
yielding slope ($p_1$) and intercept ($p_2$) as features. The slope provides an estimate of the rate of capacity loss during the chosen interval, while the intercept reflects the baseline capacity at the beginning of the fit window. Although higher-order polynomial fits were considered, they did not yield significant performance improvements in subsequent predictive modeling.

This comprehensive feature set targets both global and localized aspects of cell aging, leveraging repeated high-resolution measurements of voltage, capacity, and resistance. By combining the established framework of Severson \emph{et al.}~\cite{severson2019data} with supplemental partial-discharge descriptors from DCIR checkups, richer insights into the diverse pathways of battery degradation are enabled.

\subsection{Model selection and training}
To simultaneously perform model fitting and feature selection, a regularization-based linear regression approach was adopted. A linear model of the form
\[
\hat{y} = \hat{w} X
\]
was employed, where \(\hat{y}_i\) denotes the predicted cycle life for cell \(i\), \(X_i\) is the feature vector for cell \(i\), and \(\hat{w}\) is the coefficient vector. The elastic net regression was chosen due to its effectiveness in handling highly correlated features and relatively small sample sizes, which is characteristic of battery lifetime datasets. The elastic net formulation optimizes:
\[
\hat{w} = \text{argmin}_{w} \left(\|y - Xw\|_2^2 + \lambda \left[\alpha \|w\|_1 + (1 - \alpha)\|w\|_2^2\right]\right)
\]
where \(\lambda\) is a non-negative regularization strength scalar, and \(\alpha\) balances between the L1 norm (\(\|w\|_1\)) promoting sparsity and the L2 norm (\(\|w\|_2^2\)) promoting stability and handling collinearity.

Prior to model fitting, Sequential Feature Selection (SFS) was applied to systematically reduce dimensionality from the full initial feature set, which includes resistance-based DCIR features, voltage-capacity derived \(\Delta Q(V)\) statistics, and linear fit parameters. The optimal number of features was tuned separately for each cross-validation (CV) scheme described above (\emph{leave one triplicate out, LOTO}; \emph{leave one operating condition out, LOOCO}; and \emph{leave one manufacturer out, LOMO}), with an imposed maximum of 20 features for the \emph{LOTO} CV and 14 features for the \emph{LOOCO} and \emph{LOMO} schemes to mitigate potential overfitting. Optimum number of features was determined to be 20 for \emph{LOTO} and 12 for \emph{LOOCO} and 10 for \emph{LOMO}.

Following feature selection, Principal Component Analysis (PCA) was performed on the reduced feature set to capture variance efficiently and further reduce multi-collinearity among selected features. The elastic net model was subsequently trained using the principal components extracted from PCA.

Hyperparameter optimization of the elastic net (regularization strength \(\lambda\) and mixing parameter \(\alpha\)) was conducted via nested cross-validation to avoid data leakage, using only training folds within each CV scheme. Mean absolute errors (MAE) were computed across the held-out test folds for each CV scheme to evaluate predictive accuracy and generalizability. This approach balances model interpretability and robustness, making optimal use of the rich dataset provided by DCIR-based cycling data and ensuring a fair and rigorous assessment of the model’s ability to generalize across various manufacturers and cycling conditions.

To interpret feature importances in the original input space, we applied model-agnostic SHAP analysis to a custom-wrapped pipeline comprising sequential feature selection (SFS), principal component analysis (PCA), and ElasticNet regression. By exposing a .predict() interface on the full pipeline and supplying untransformed input data to a KernelExplainer, SHAP perturbations and attributions were computed with respect to the preprocessed (raw) feature set rather than post-PCA components.

\section{Data Availability}
The data generated and analyzed during this study, including raw cycling data, cell metadata, summary diagnostic metrics, and Neware cycling protocol files, are available at the Open Science Framework repository: [link redacted for journal review. Interested readers should contact authors for early access]

\section*{Acknowledgements}
The authors acknowledge Jacob Lokshin, Brian Wee, Yunyun Wang, Brenda Miao, Dan Zhao and Joshua Douglas for their invaluable contributions and support throughout this investigation.

\newpage

\bibliographystyle{plainnat}
\bibliography{main}

\newpage
\appendix

\section{Model Validation}
\label{sec:cv}

\begin{figure}[ht!]
    \centering
    \begin{subfigure}[b]{0.40\columnwidth}
        \includegraphics[width=\linewidth]{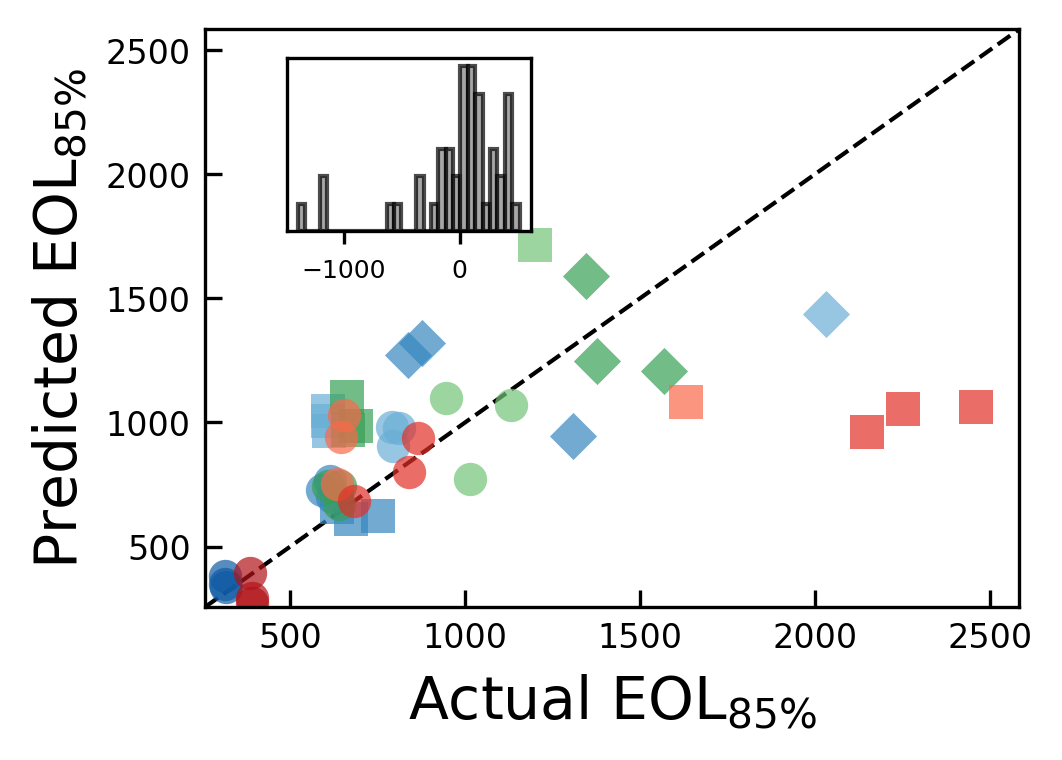}
        \caption{}
        \label{fig:kfold_dqv_parity}
    \end{subfigure}
    \hspace{0.2cm}
    \begin{subfigure}[b]{0.40\columnwidth}
        \includegraphics[width=\linewidth]{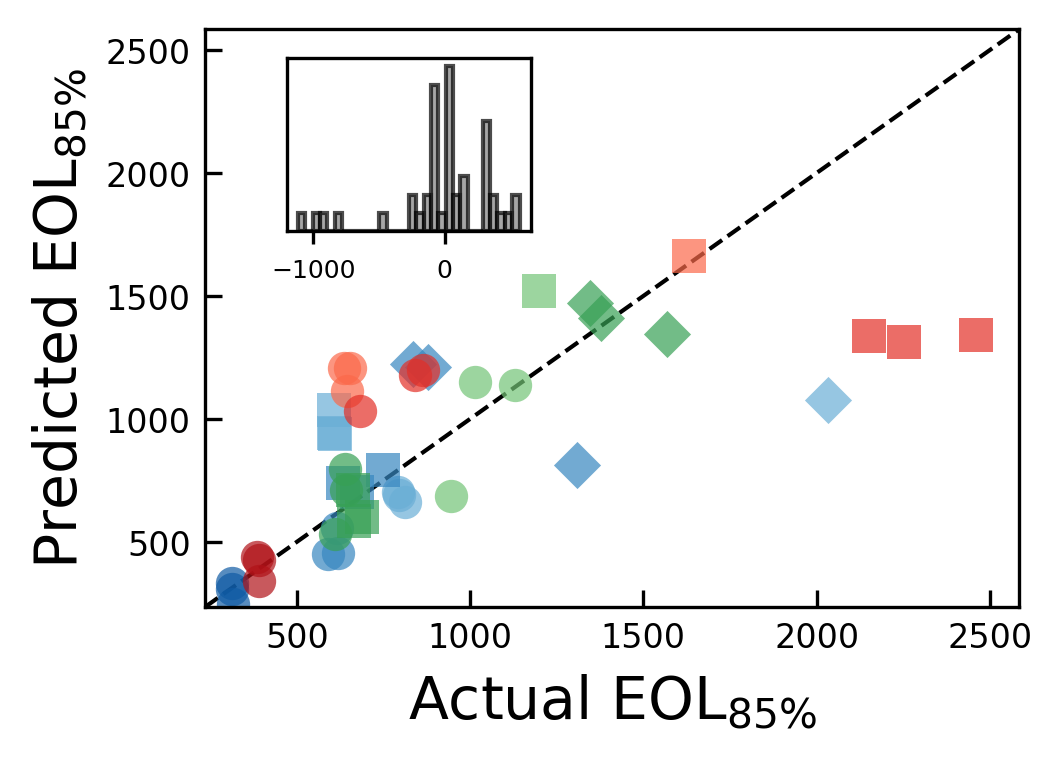}
        \caption{}
        \label{fig:kfold_nondcir_parity}
    \end{subfigure}
    
    \begin{subfigure}[b]{0.40\columnwidth}
        \includegraphics[width=\linewidth]{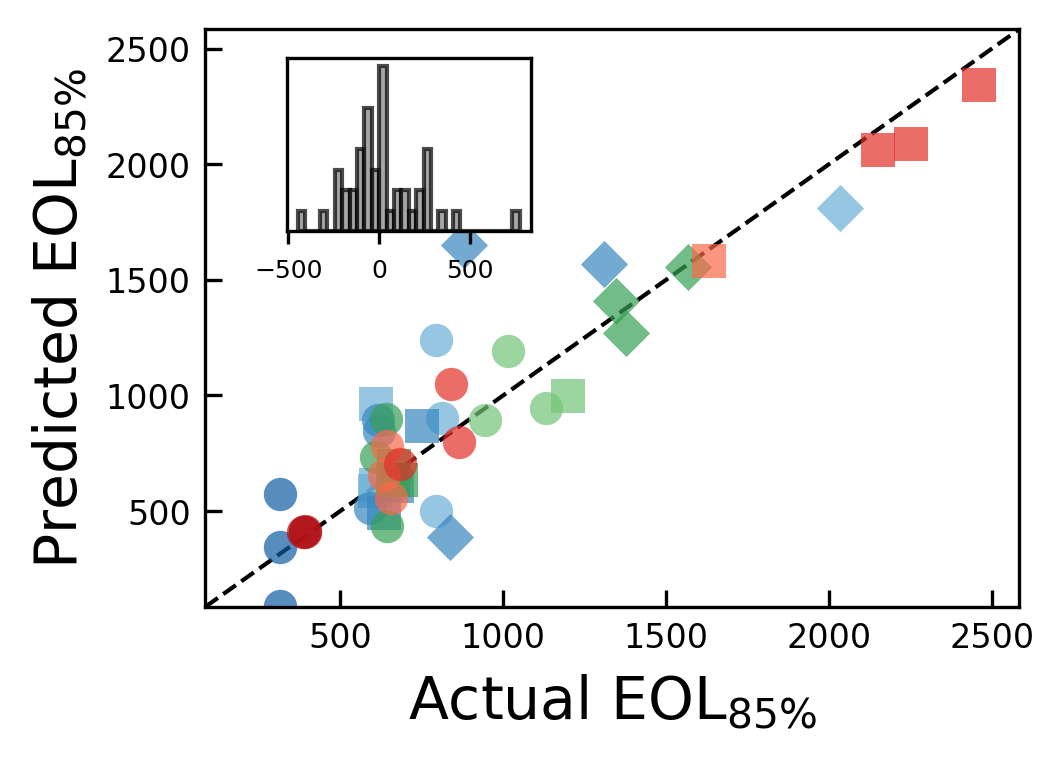}
        \caption{}
        \label{fig:kfold_dcir_dqv_parity}
    \end{subfigure}
    \hspace{0.2cm}
    \begin{subfigure}[b]{0.40\columnwidth}
        \includegraphics[width=\linewidth]{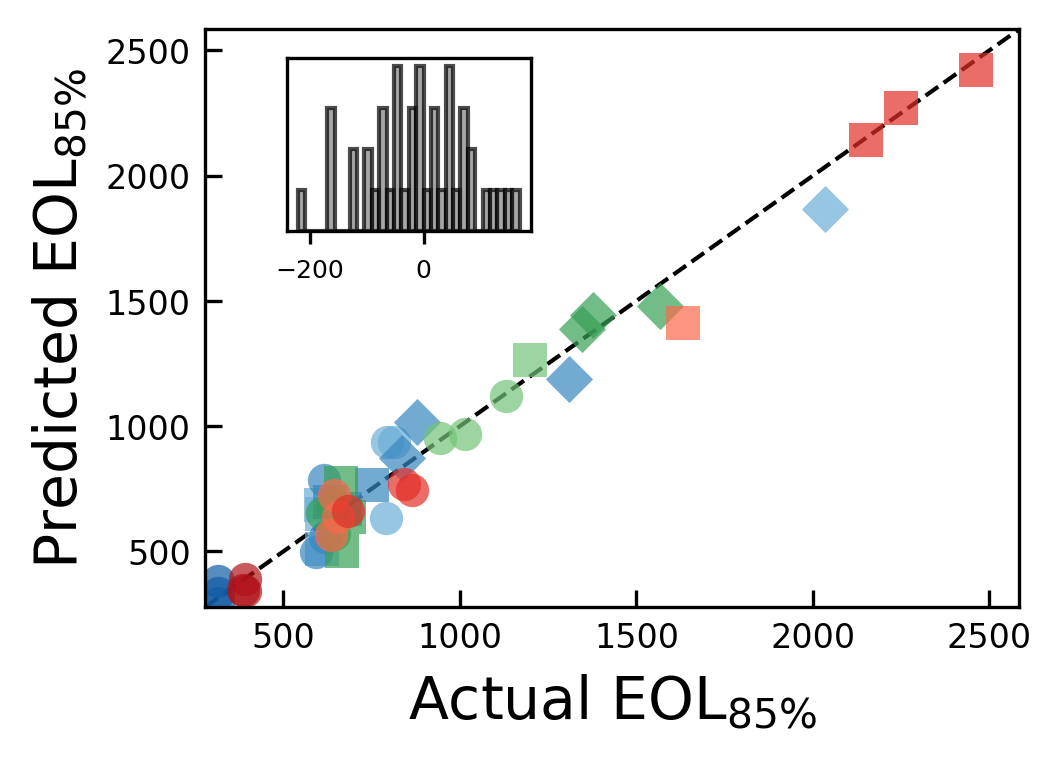}
        \caption{}
        \label{fig:kfold_all_dcir_parity}
    \end{subfigure}
    
    \begin{subfigure}[b]{0.40\columnwidth}
        \includegraphics[width=\linewidth]{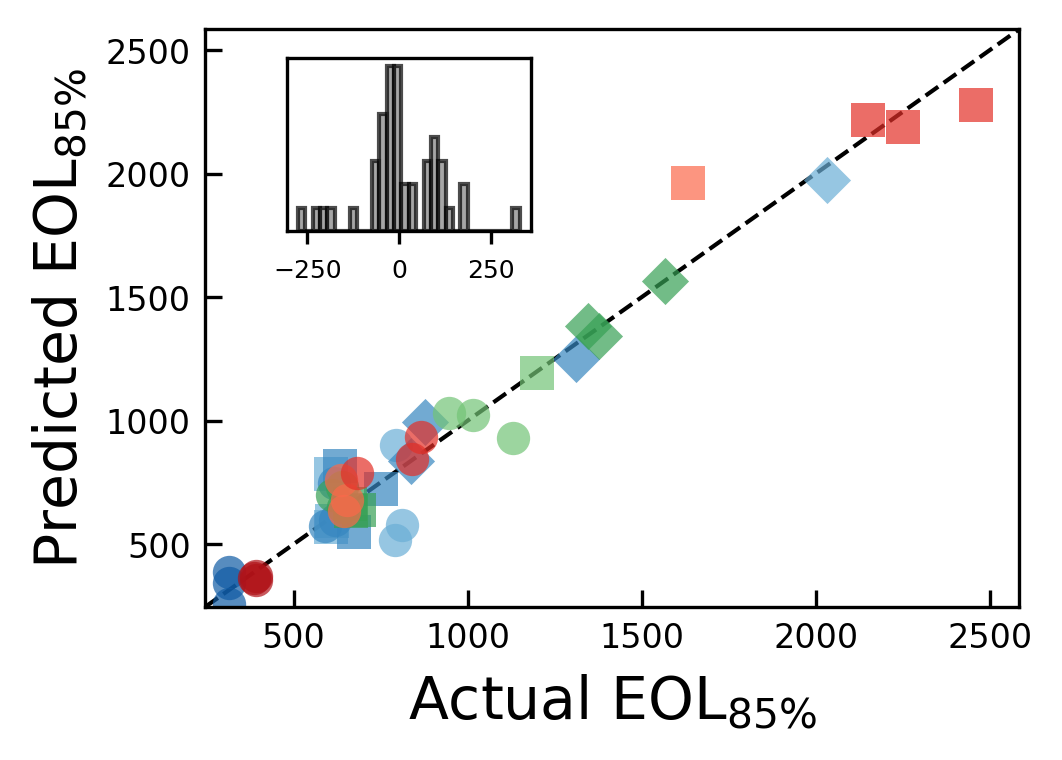}
        \caption{}
        \label{fig:kfold_all_parity}
    \end{subfigure}
    
    \caption{Parity plots showing the 5fold CV prediction accuracy of models trained on various feature sets: (a) the original six feature $\Delta$Q(V) set from Severson et al, (b) the original six feature $\Delta$Q(V) set and linear fit of the lifetime curve, (c) the novel augmented DCIR $\Delta$Q(V) feature set, (d) the augmented DCIR $\Delta$Q(V) and pulse resistances, and (e) the full combined feature set as described in the manuscript.}
    \label{fig:five_panel_layout}
\end{figure}

In addition to the "Leave One Triplicate Out" cross validation strategy employed in the manuscript, we perform traditional 5 fold cross-validation for direct comparison. The following parity plots and table are equivalent analyses to that presented in the main text, but with sklearn's standard KFold cross validator with 5 splits (80/20 training/testing split). Shown in the parity plots of figs. \ref{fig:kfold_dqv_parity}-\ref{fig:kfold_all_parity} are the compiled test set predictions of all 5 test folds. The reported MAE in table \ref{tab:sfs_features} are the averaged values of all test folds.

\begin{table}[h]
\centering
\resizebox{0.8\textwidth}{!}{
\begin{tabular}{lccc}
    \toprule
    Features & EOL$_{80}$ & EOL$_{85}$ & EOL$_{90}$ \\
    \midrule
    $\Delta Q(V)_{101-1}$ & 359 & 259 & 134 \\
    Non-DCIR features & 366 & 256 & 122 \\
    DCIR-based features & 202 & 109 & 40 \\
    All features & 163 & 145 & 46 \\
    \bottomrule
\end{tabular}
}
\caption{Model performance for Elastic Net architecture using a standard 5-fold cross validation technique.}
\end{table}

\section{Feature selection}
The feature space was dimensionally reduced using Sequential Feature Selection (SFS). For each cross-validation scheme considered in this work, Table \ref{tab:sfs_features} shows the number of features retained and that subsequently underwent Principle Component Analysis (PCA). 

\begin{table}[h]
\centering
\resizebox{0.8\textwidth}{!}{
\begin{tabular}{lc}
    \toprule
    CV scheme & Feature space dimension \\
    \midrule
    Leave one triplicate out & 20 \\
    Leave one operating condition out & 12 \\
    Leave one manufacturer out & 10 \\
    \bottomrule
\end{tabular}
}
\caption{Outcome of sequential feature selection showing final feature space dimensions for each cross-validation (CV) scheme.}
\label{tab:sfs_features}
\end{table}

\clearpage

\section{Data quality and end-of-life labeling}
For the results presented in this work, end-of-life was defined as the number of cycles required to reach 85\% State of Health (SOH). The following figures provide an overview of the individual lifetimes for each cell with the end-of-life point indicated by a vertical line at the cycle considered to be end-of-life. Note that a few of the cells (e.g., Sam50E at 25 degrees C, 3.3 - 4.2 V) did not fully reach 85\%, and a linear fit of the last 100 cycles was used to extrapolate to 85\% SOH. For these cells, the EOL label should be taken with caution and may be overestimating the actual EOL if the cell were to "nosedive" and experience rapid loss in capacity at some point in the extrapolation window. That said, due to the known challenges with data scarcity, we believe linear extrapolation provides a reasonable estimation of the EOL for these cells and the benefits of the additional data outweigh possible risks of overestimation from the methodologies used.

\begin{figure}[h]
    \centering
    \includegraphics[width=0.9\textwidth]{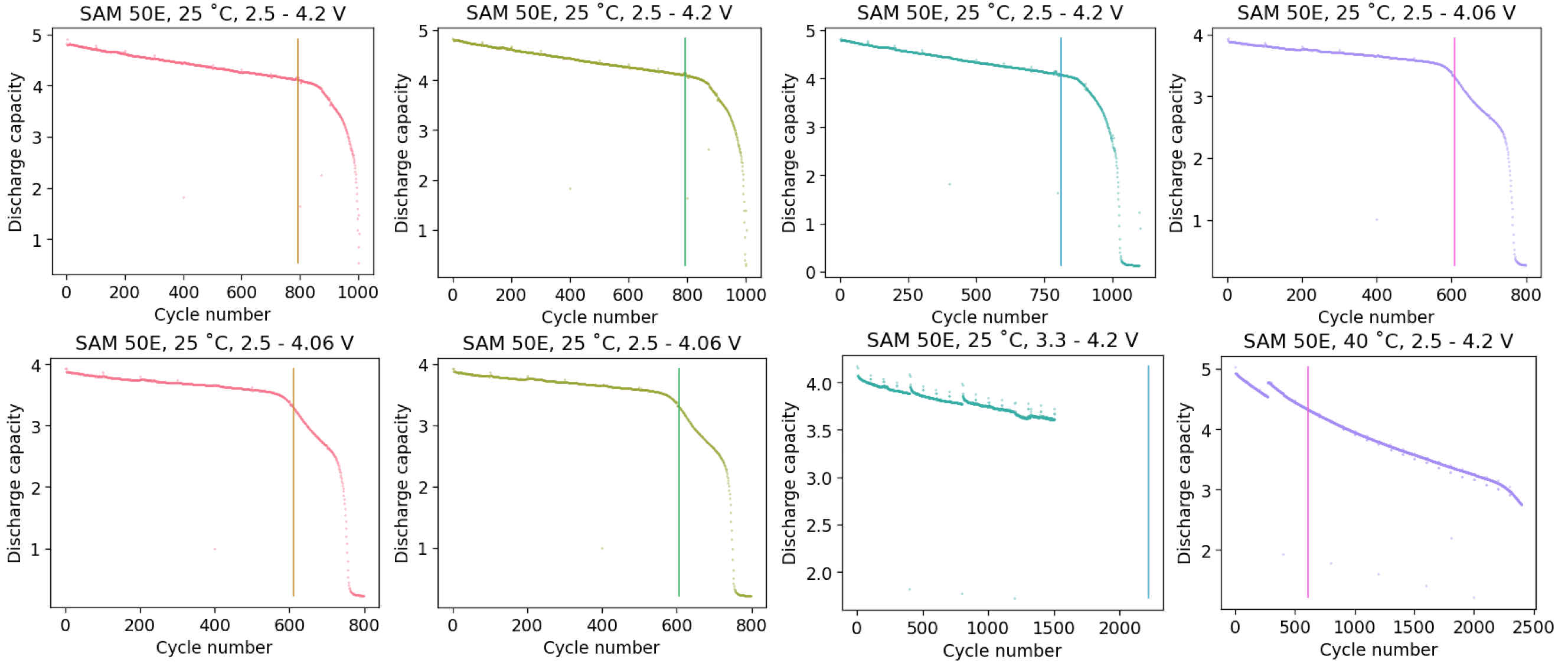}
    \caption{Overview of individual cell lifetimes with labeled end-of-life denoted with vertical line.}
    \label{fig:lifetimes_0}
\end{figure}

\begin{figure}[h]
    \centering
    \includegraphics[width=0.9\textwidth]{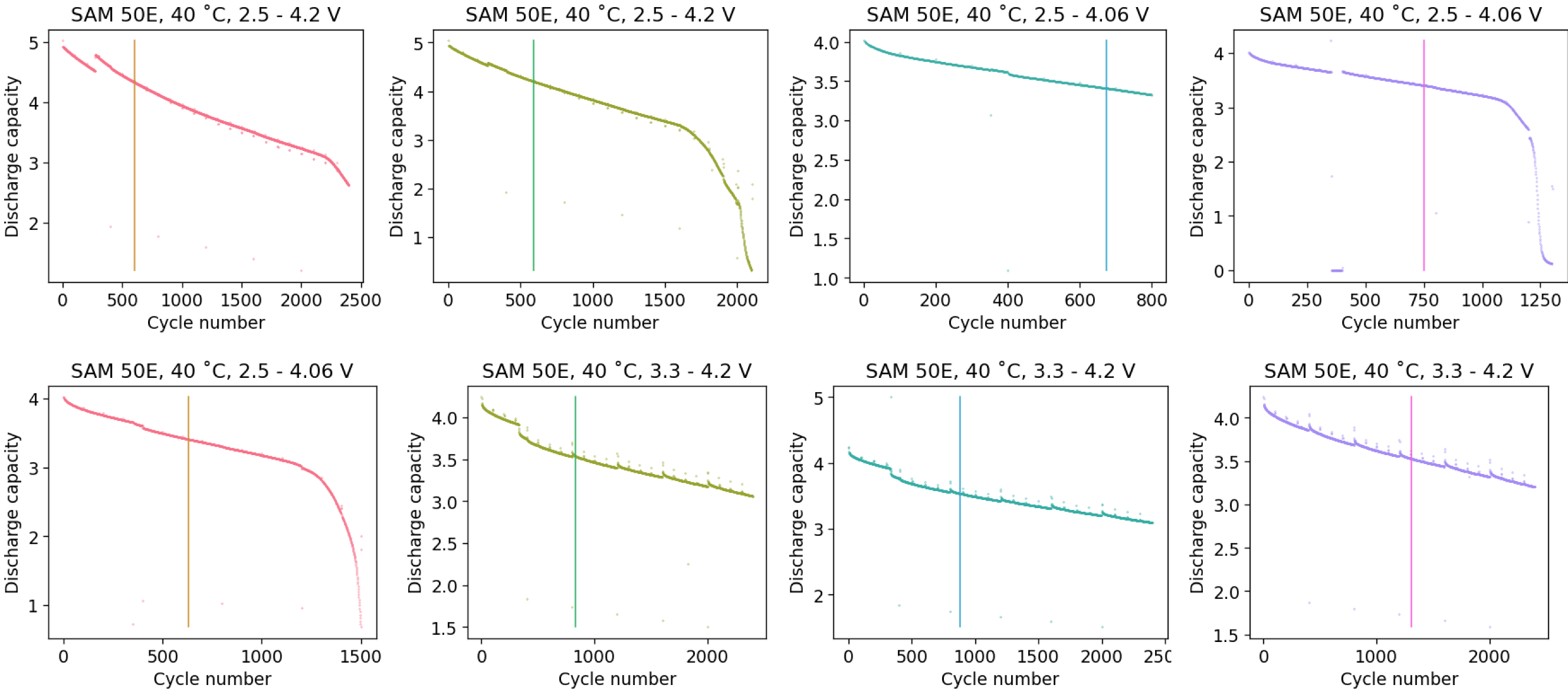}
    \caption{Overview of individual cell lifetimes with labeled end-of-life denoted with vertical line.}
    \label{fig:lifetimes_1}
\end{figure}

\begin{figure}[h]
    \centering
    \includegraphics[width=0.9\textwidth]{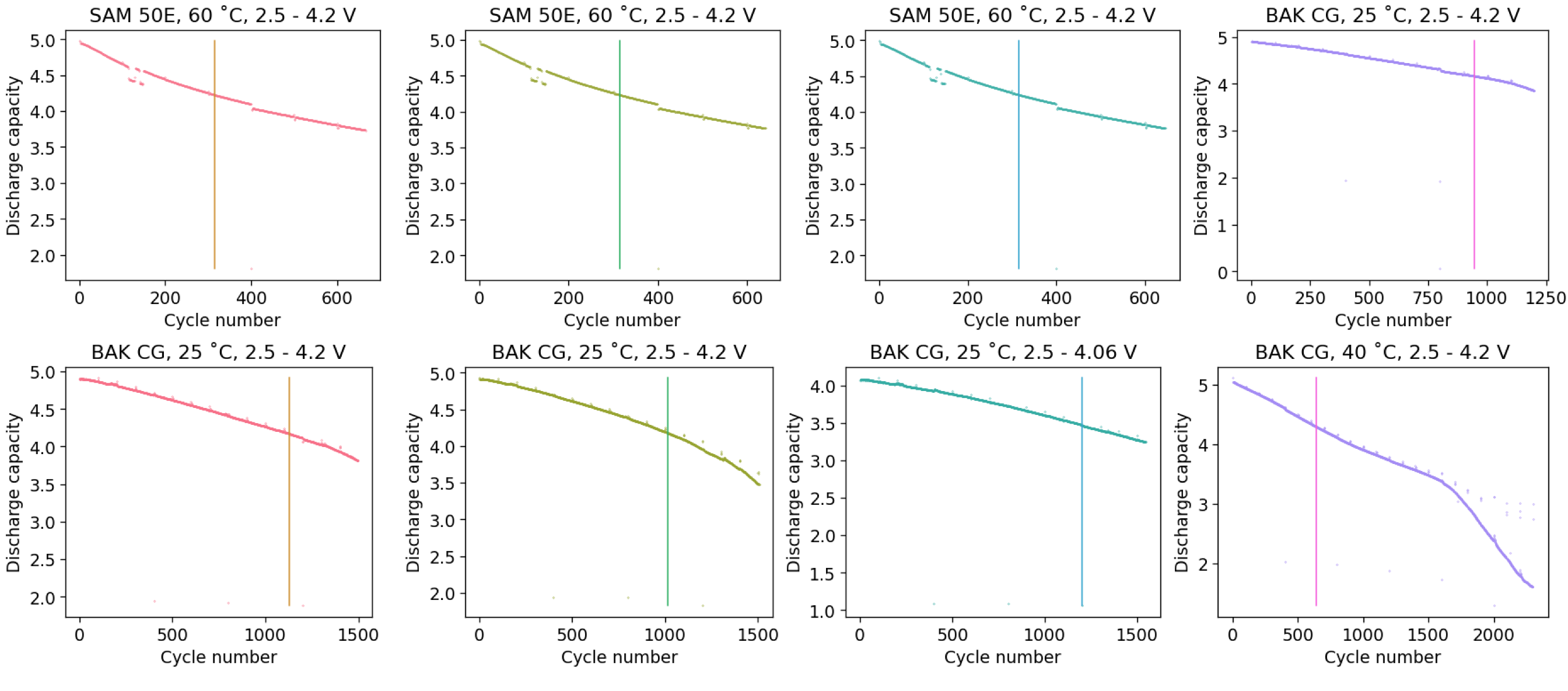}
    \caption{Overview of individual cell lifetimes with labeled end-of-life denoted with vertical line.}
    \label{fig:lifetimes_2}
\end{figure}

\begin{figure}[h]
    \centering
    \includegraphics[width=0.9\textwidth]{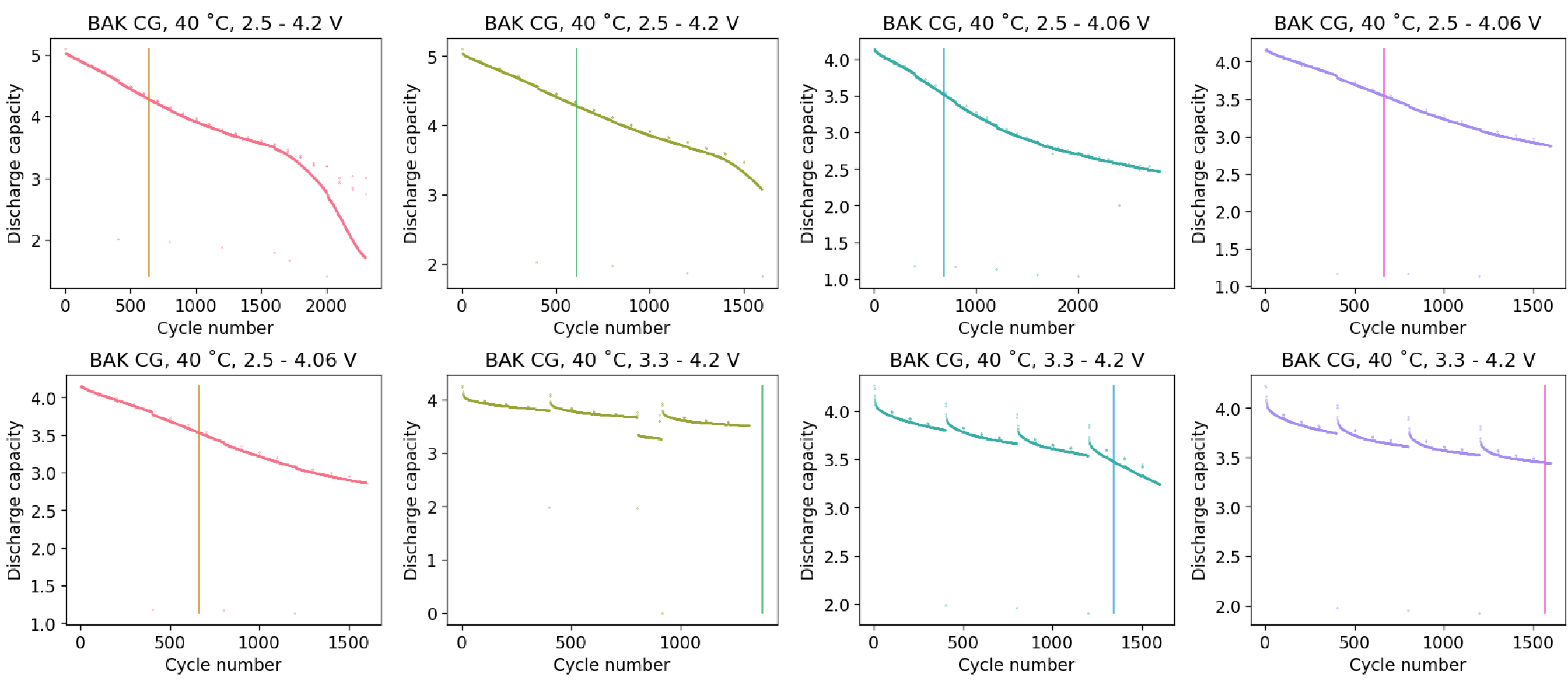}
    \caption{Overview of individual cell lifetimes with labeled end-of-life denoted with vertical line.}
    \label{fig:lifetimes_3}
\end{figure}

\begin{figure}[h]
    \centering
    \includegraphics[width=0.9\textwidth]{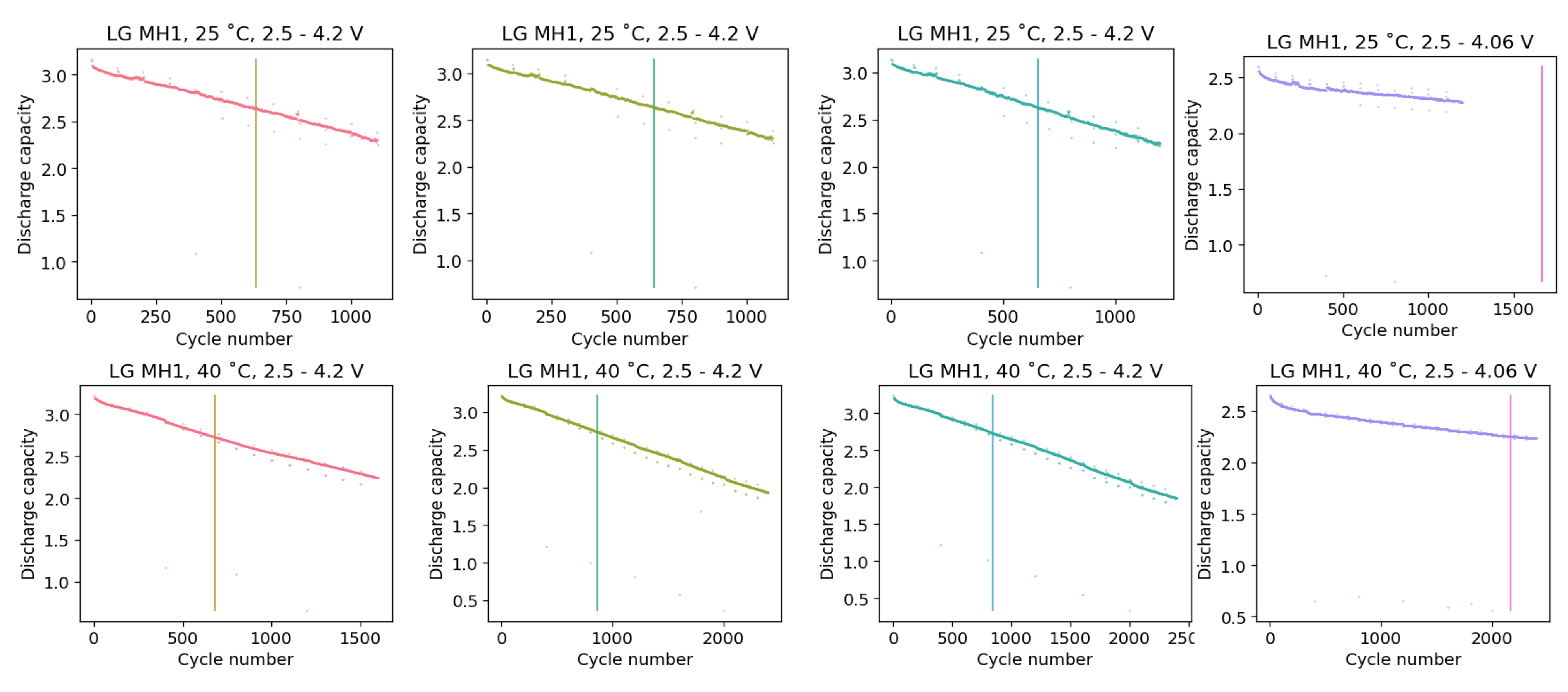}
    \caption{Overview of individual cell lifetimes with labeled end-of-life denoted with vertical line.}
    \label{fig:lifetimes_4}
\end{figure}

\begin{figure}[h]
    \centering
    \includegraphics[width=0.9\textwidth]{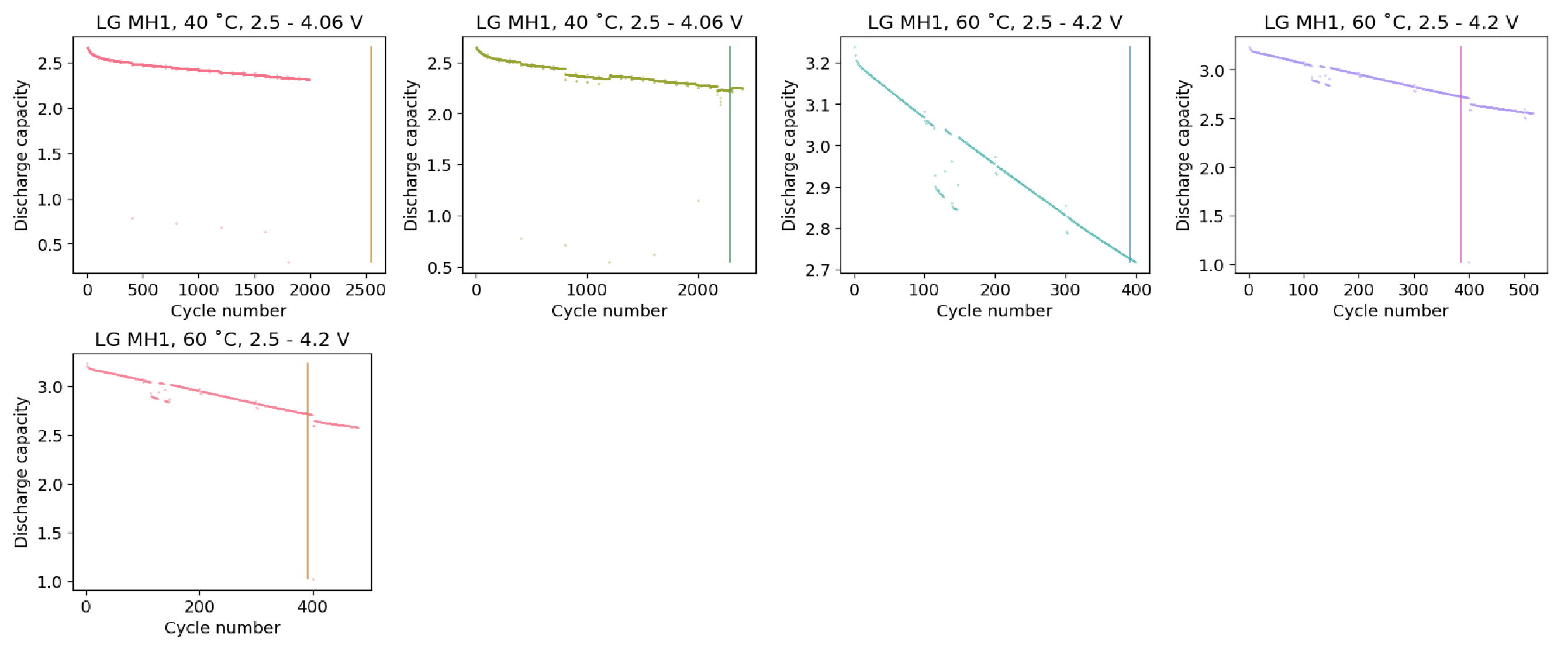}
    \caption{Overview of individual cell lifetimes with labeled end-of-life denoted with vertical line.}
    \label{fig:lifetimes_5}
\end{figure}

\clearpage

\section{Feature correlation}
\begin{figure}[h]
    \centering
    \includegraphics[width=0.8\textwidth]{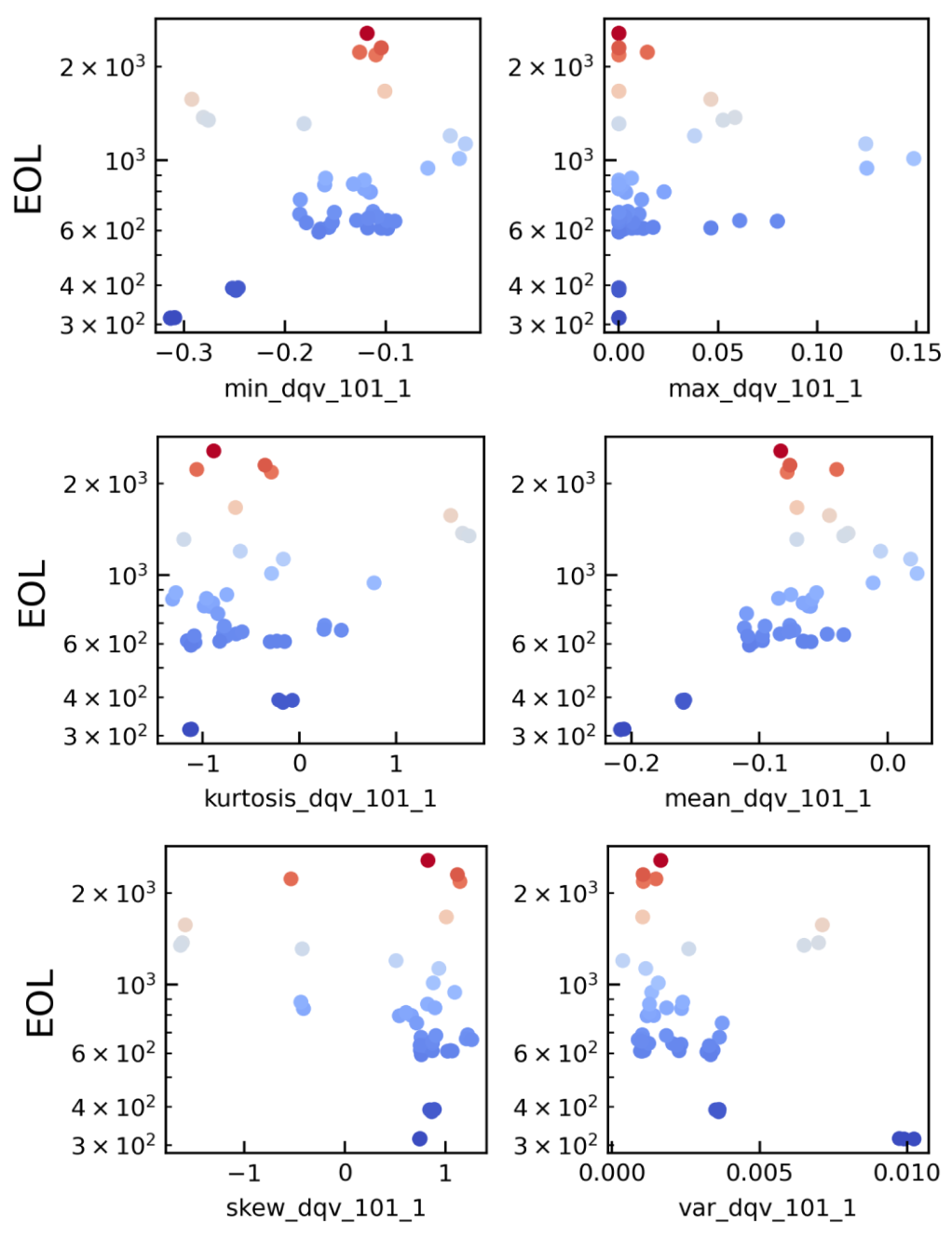}
    \caption{Feature correlation of the traditional Q(V) feature set.}
    \label{fig:pairplots_dqv}
\end{figure}

\begin{figure}[h]
    \centering
    \includegraphics[width=0.8\textwidth]{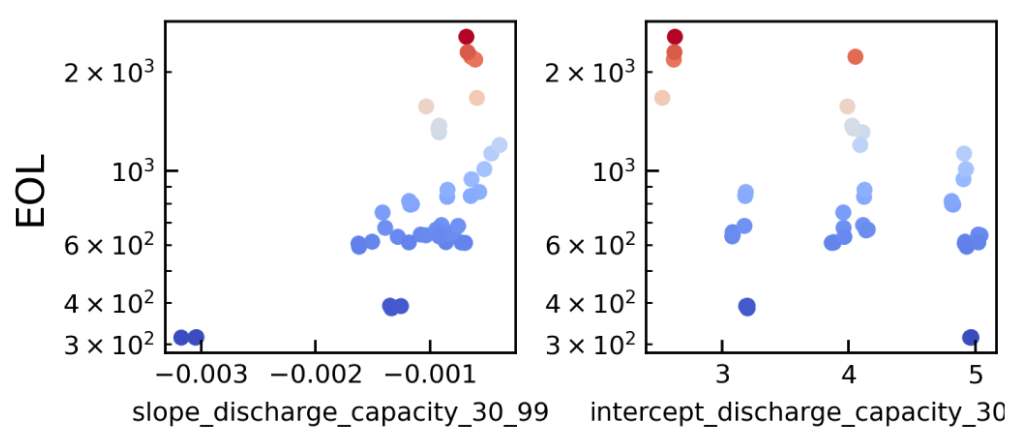}
    \caption{Feature correlation of the p1p2 feature set.}
    \label{fig:pairplots_p1p2}
\end{figure}

\begin{figure}[h]
    \centering
    \includegraphics[width=0.8\textwidth]{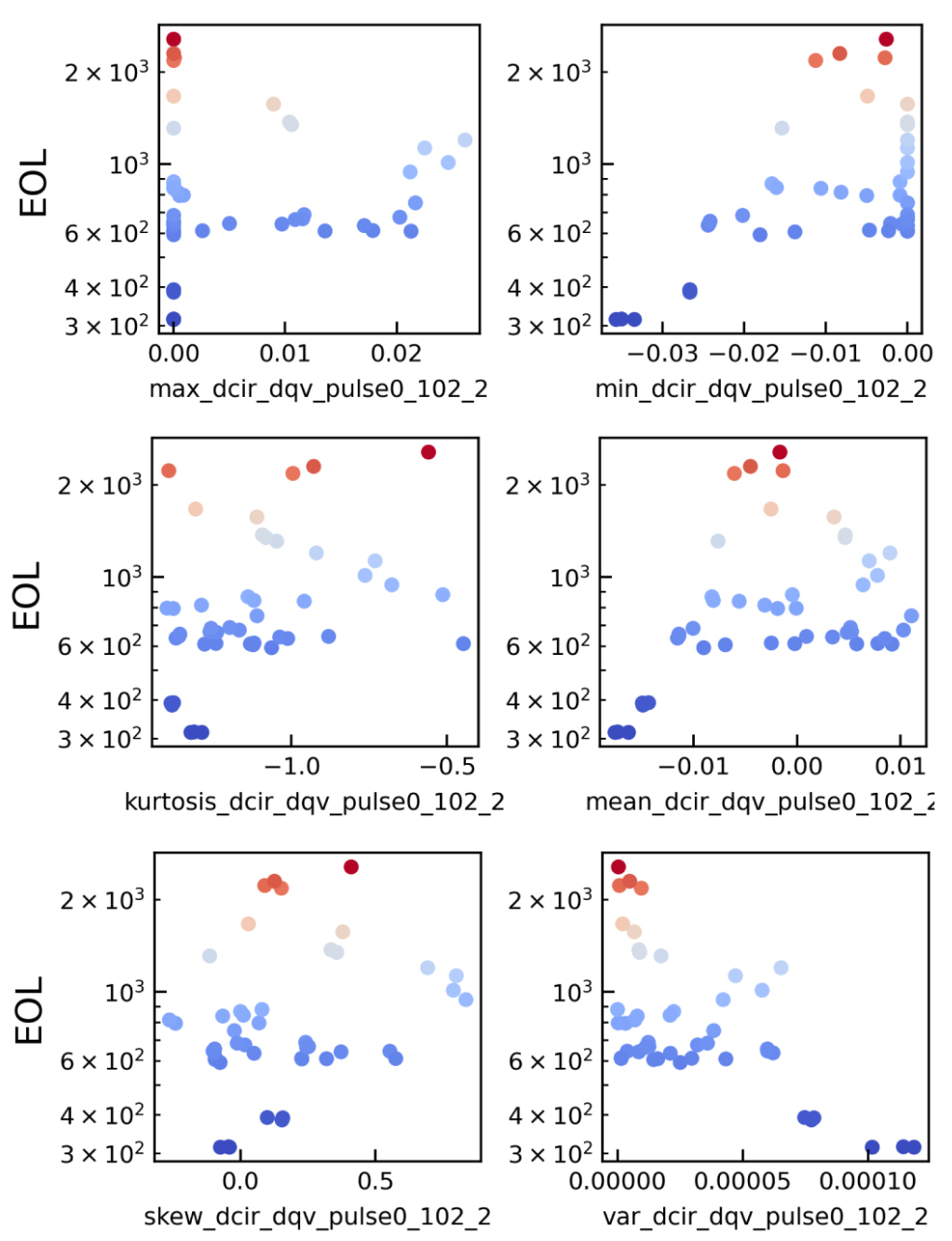}
    \caption{Feature correlation of the DCIR Q(V) feature set for pulse0.}
    \label{fig:pairplots_dqv_pulse0}
\end{figure}

\begin{figure}[h]
    \centering
    \includegraphics[width=0.8\textwidth]{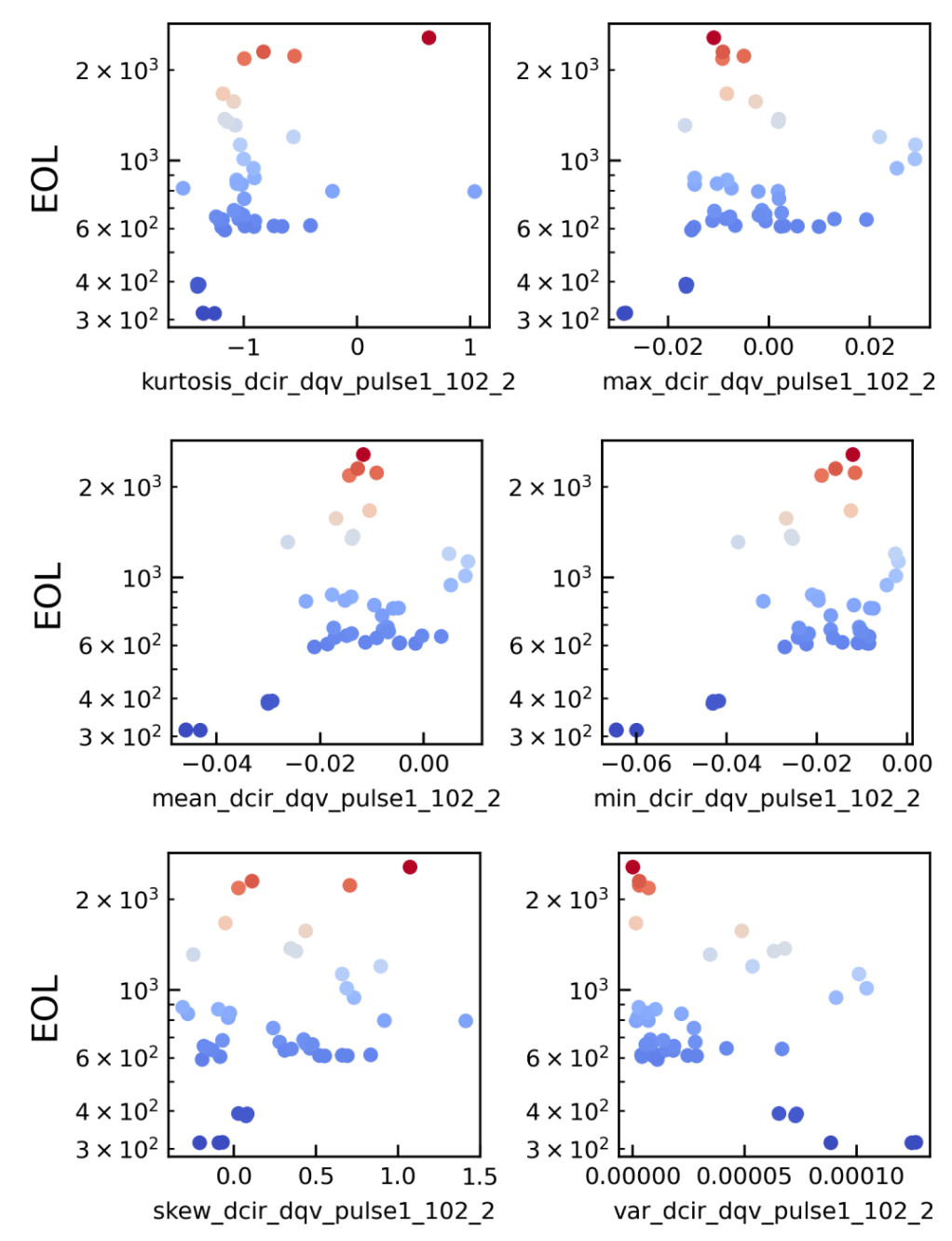}
    \caption{Feature correlation of the DCIR Q(V) feature set for pulse1.}
    \label{fig:pairplots_dqv_pulse1}
\end{figure}

\begin{figure}[h]
    \centering
    \includegraphics[width=0.8\textwidth]{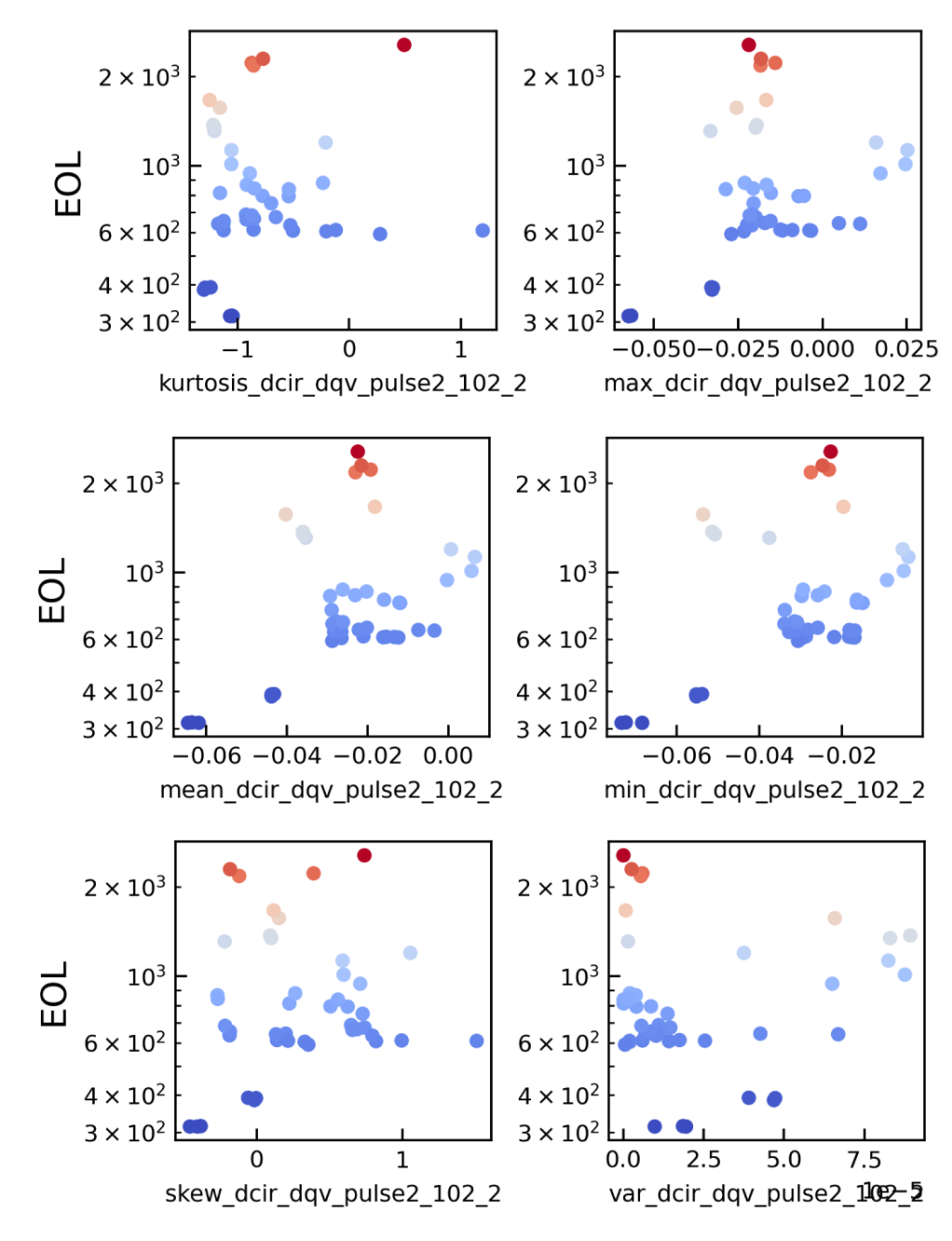}
    \caption{Feature correlation of the DCIR Q(V) feature set for pulse2.}
    \label{fig:pairplots_dqv_pulse2}
\end{figure}

\begin{figure}[h]
    \centering
    \includegraphics[width=0.8\textwidth]{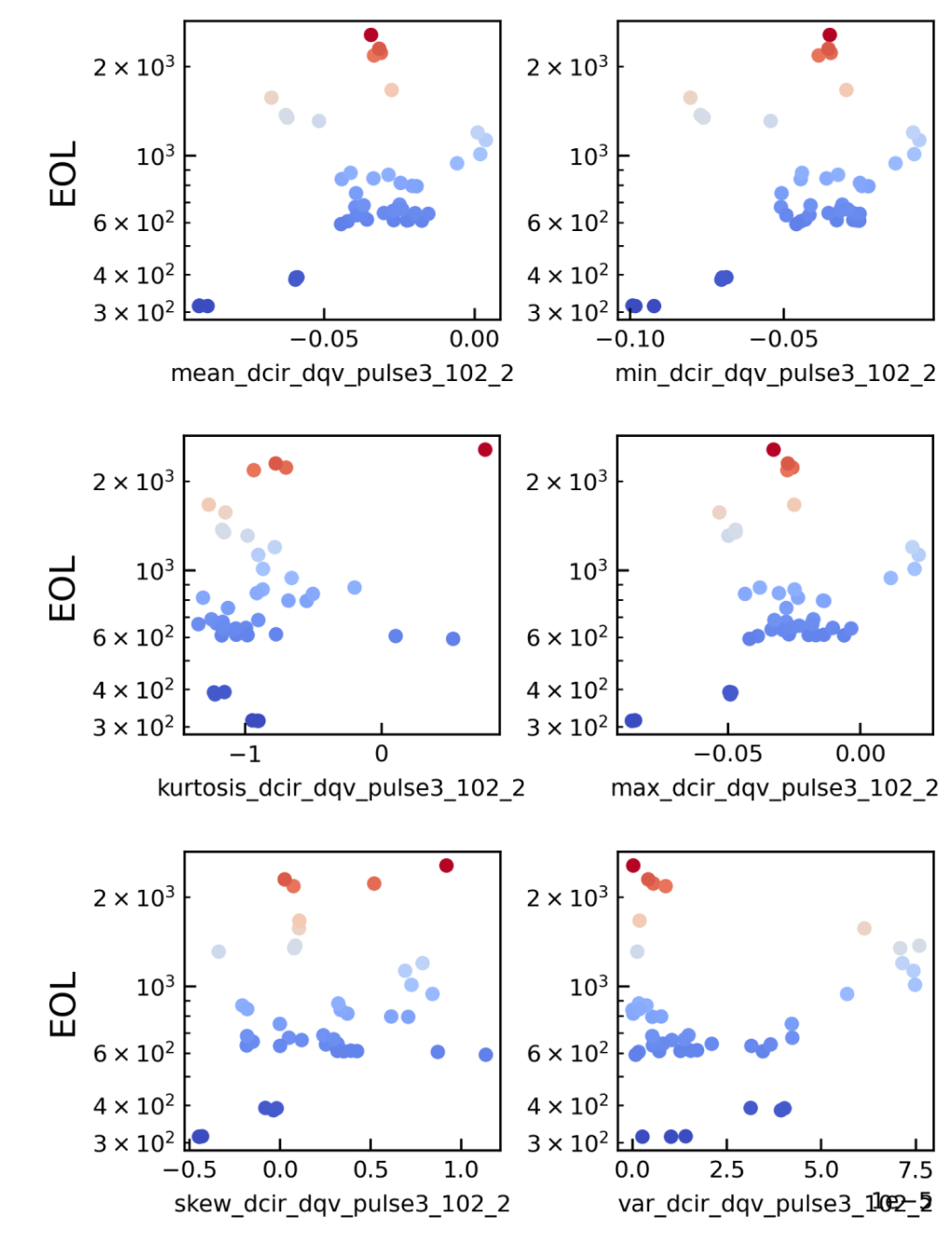}
    \caption{Feature correlation of the DCIR Q(V) feature set for pulse3.}
    \label{fig:pairplots_dqv_pulse3}
\end{figure}

\begin{figure}[h]
    \centering
    \includegraphics[width=0.8\textwidth]{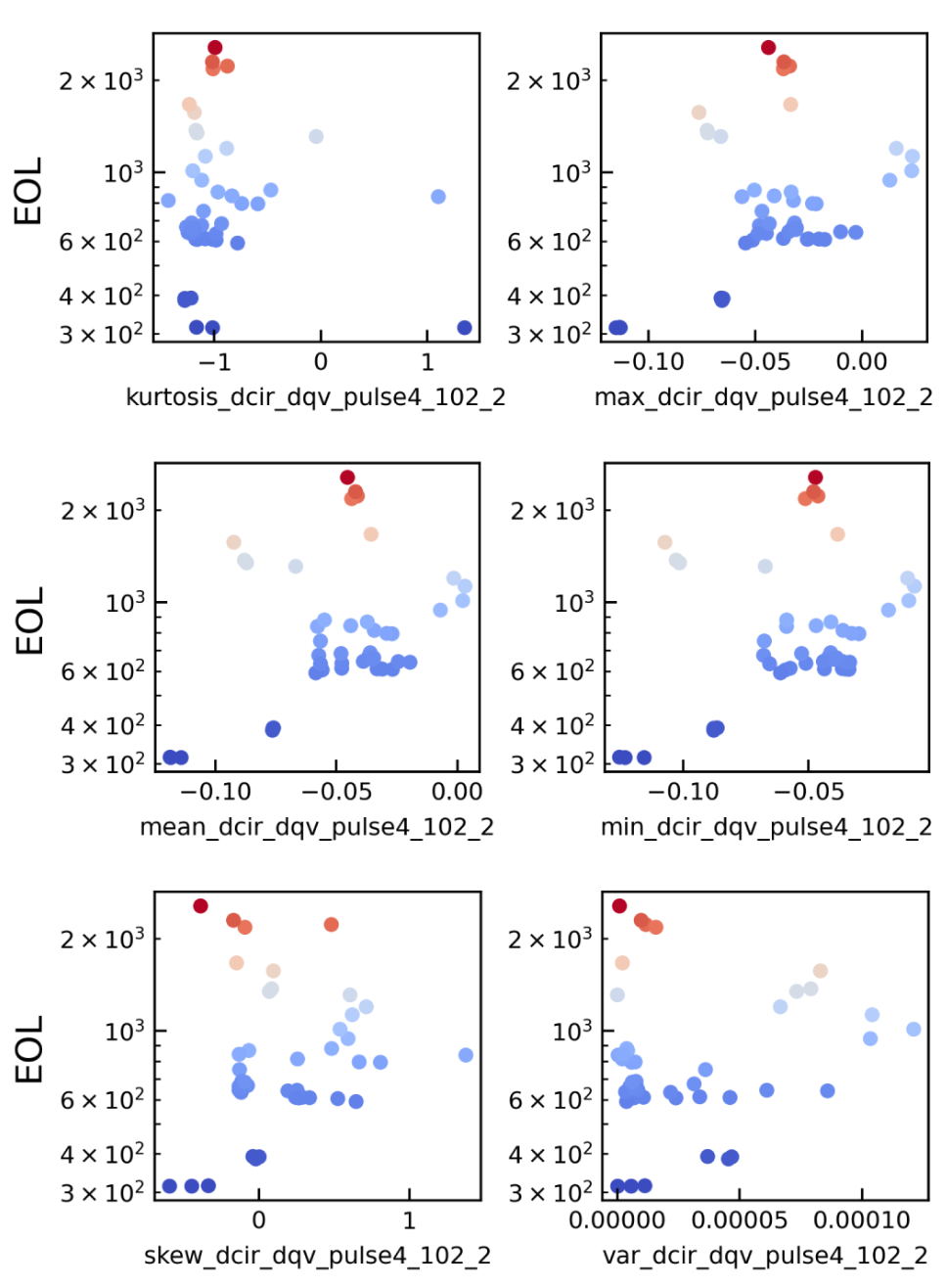}
    \caption{Feature correlation of the DCIR Q(V) feature set for pulse4.}
    \label{fig:pairplots_dqv_pulse4}
\end{figure}

\begin{figure}[h]
    \centering
    \includegraphics[width=0.8\textwidth]{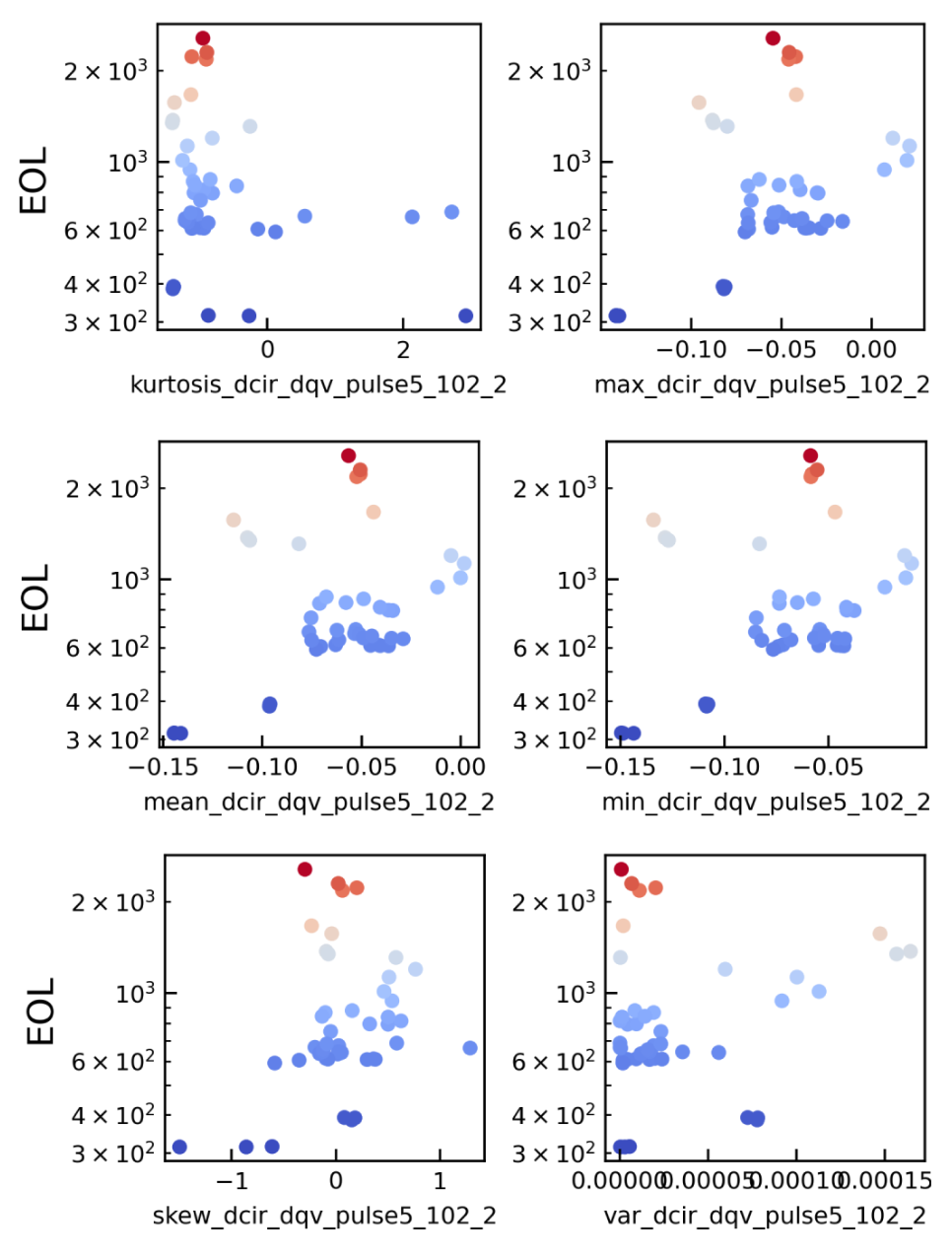}
    \caption{Feature correlation of the DCIR Q(V) feature set for pulse5.}
    \label{fig:pairplots_dqv_pulse5}
\end{figure}

\begin{figure}[h]
    \centering
    \includegraphics[width=0.8\textwidth]{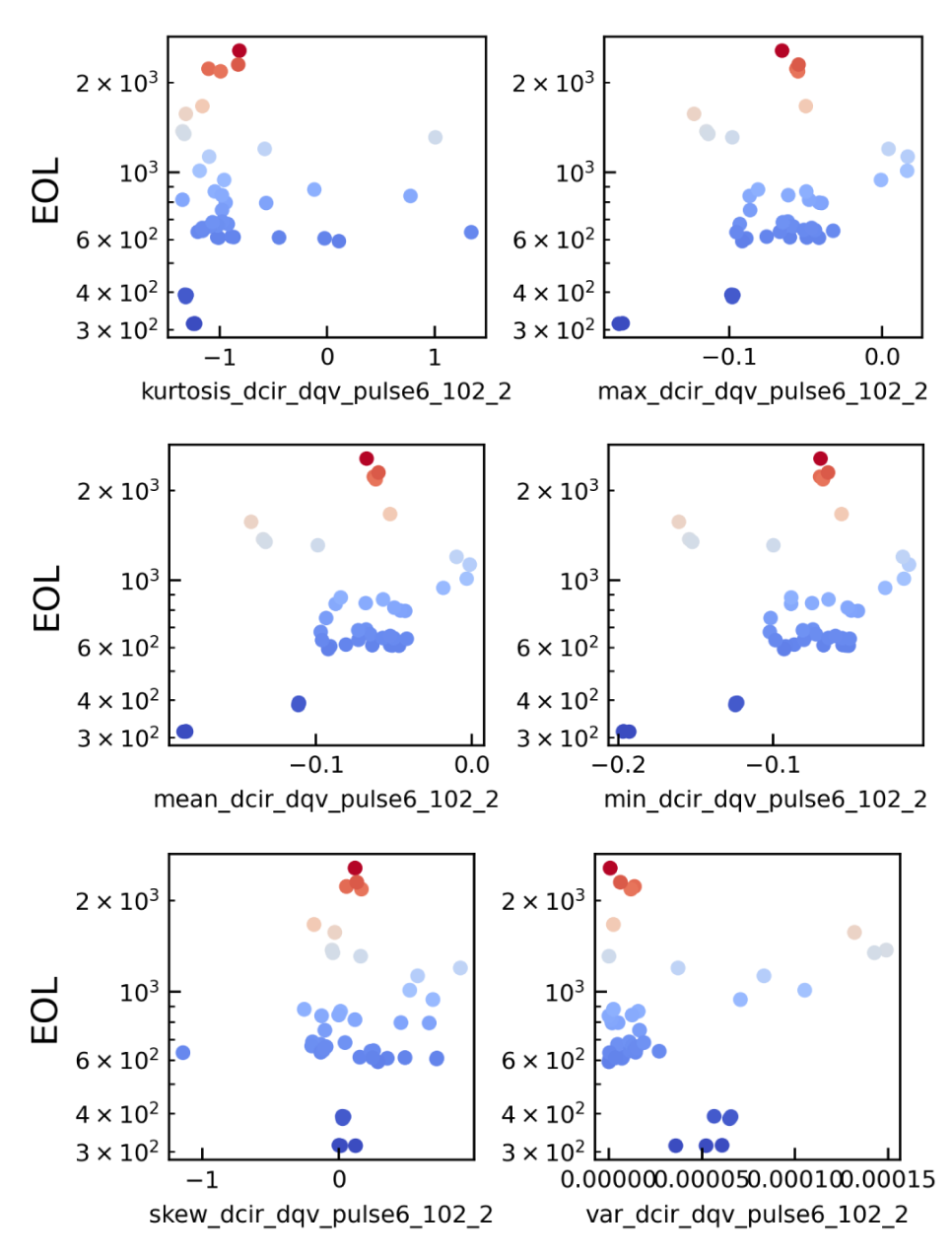}
    \caption{Feature correlation of the DCIR Q(V) feature set for pulse6.}
    \label{fig:pairplots_dqv_pulse6}
\end{figure}

\begin{figure}[h]
    \centering
    \includegraphics[width=0.8\textwidth]{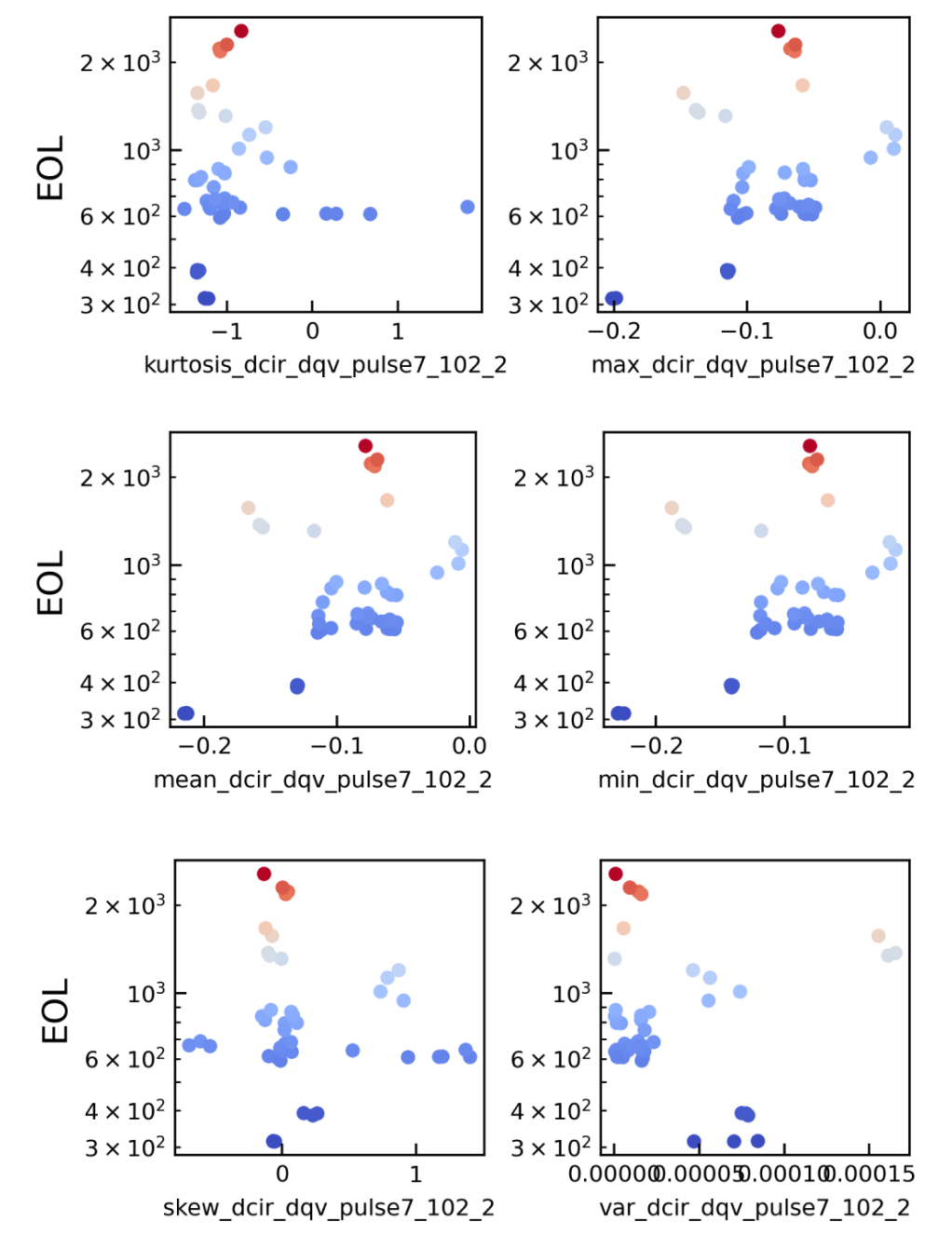}
    \caption{Feature correlation of the DCIR Q(V) feature set for pulse7.}
    \label{fig:pairplots_dqv_pulse7}
\end{figure}

\begin{figure}[h]
    \centering
    \includegraphics[width=0.8\textwidth]{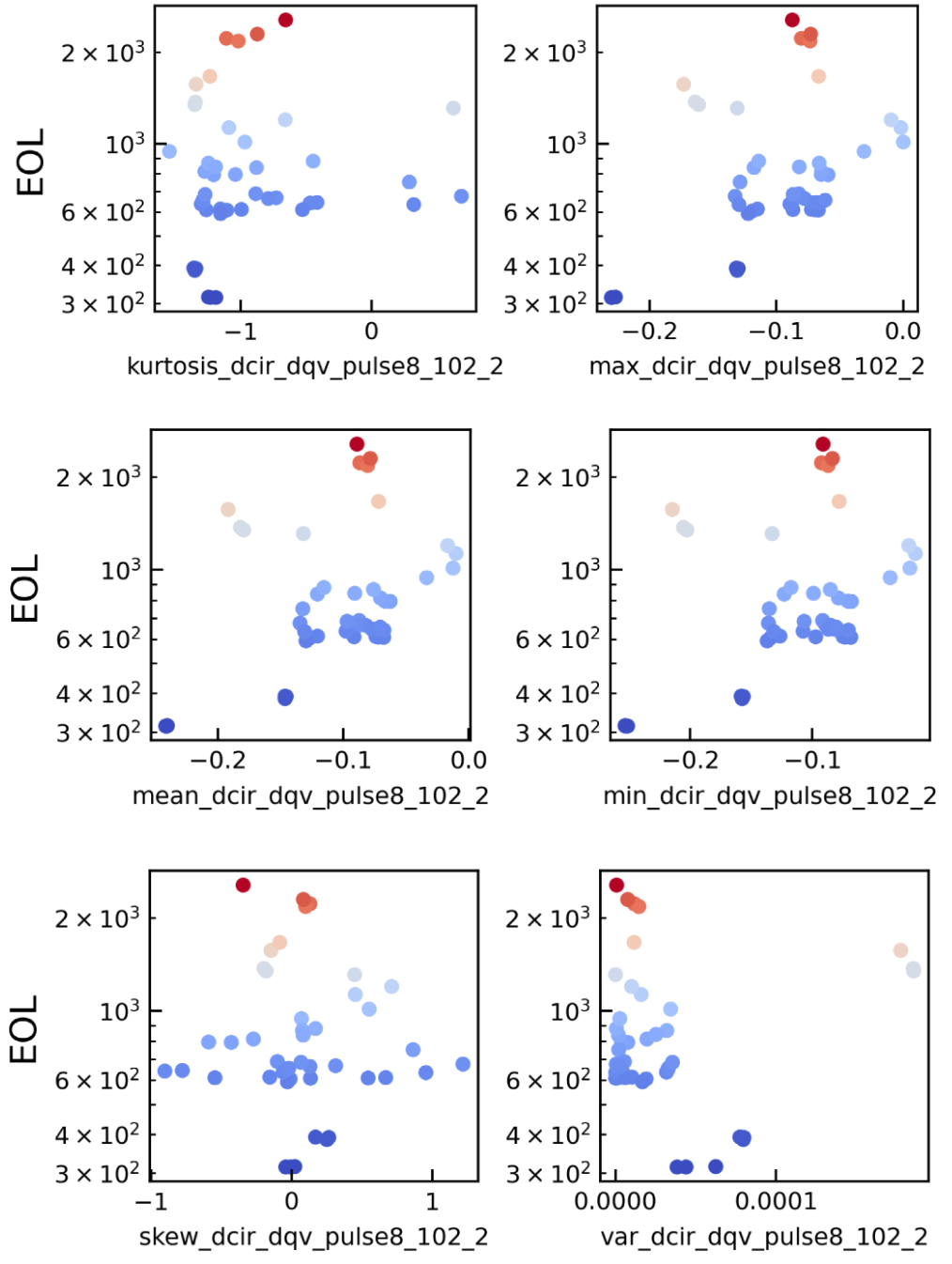}
    \caption{Feature correlation of the DCIR Q(V) feature set for pulse8.}
    \label{fig:pairplots_dqv_pulse8}
\end{figure}

\begin{figure}[h]
    \centering
    \includegraphics[width=0.8\textwidth]{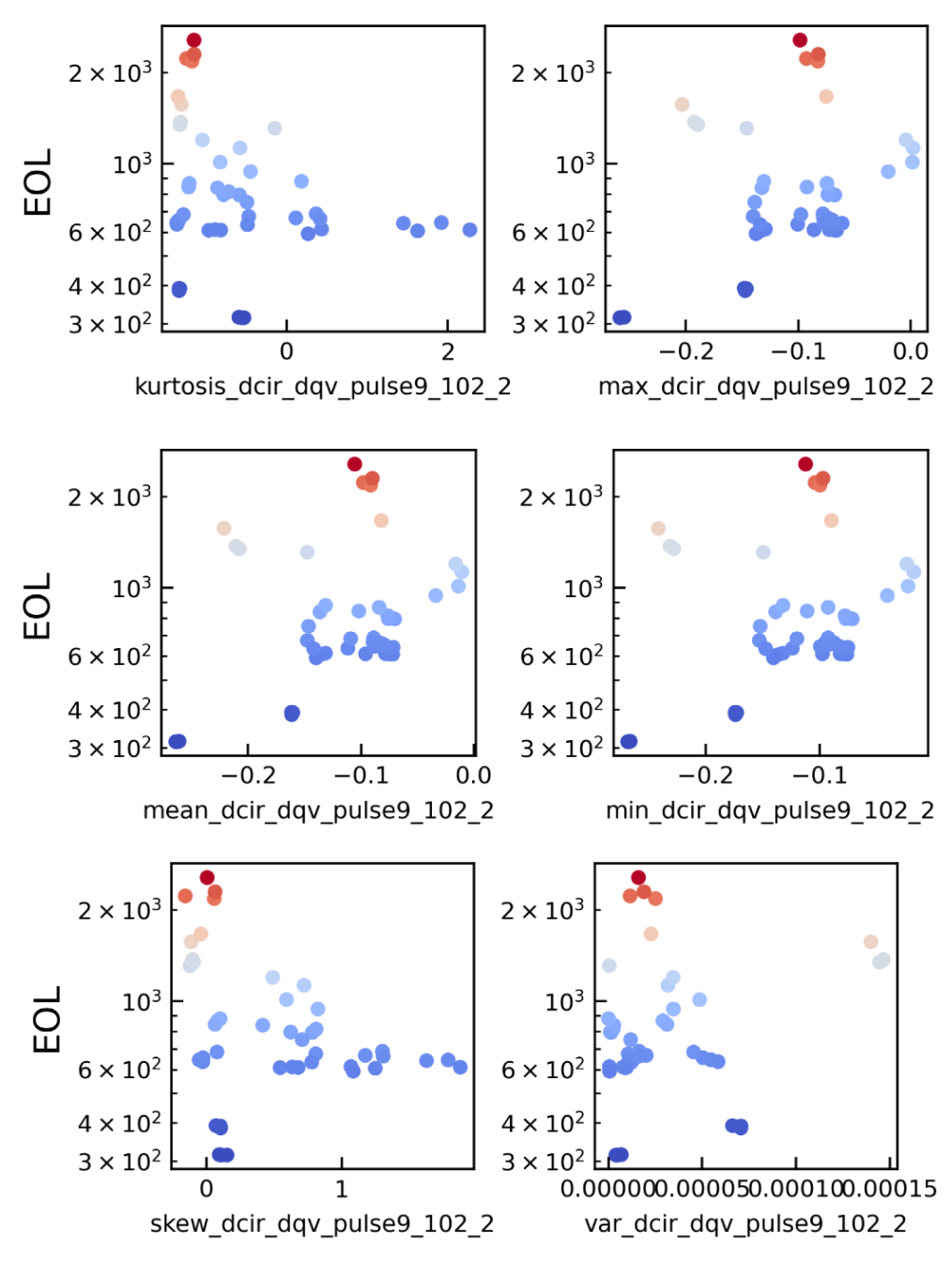}
    \caption{Feature correlation of the DCIR Q(V) feature set for pulse9.}
    \label{fig:pairplots_dqv_pulse9}
\end{figure}

\begin{figure}[h]
    \centering
    \includegraphics[width=0.8\textwidth]{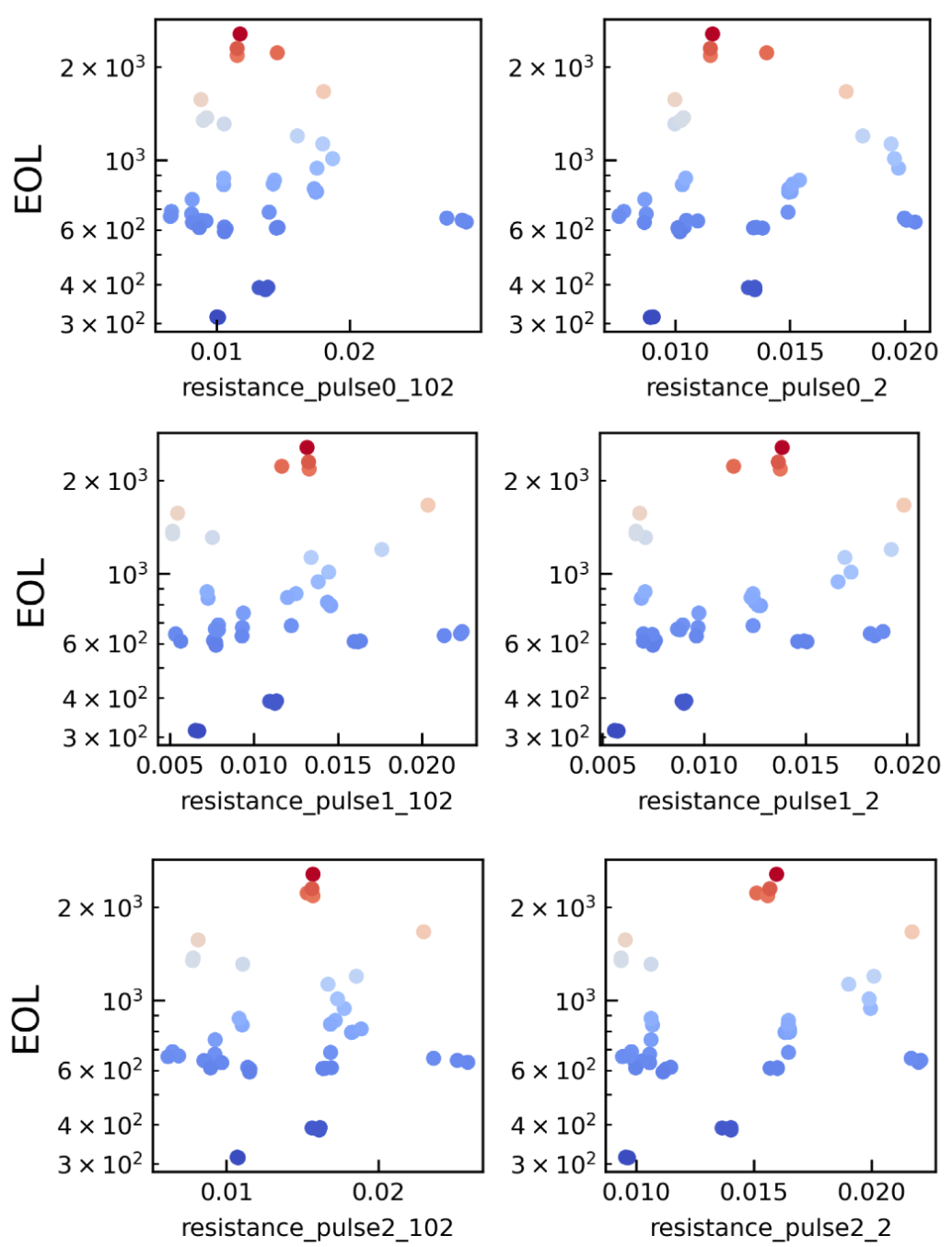}
    \caption{Feature correlation of the DCIR resistances of cycle 102 and 2 for pulse0, pulse1, and pulse2.}
    \label{fig:pairplots_res_pulse012}
\end{figure}

\begin{figure}[h]
    \centering
    \includegraphics[width=0.8\textwidth]{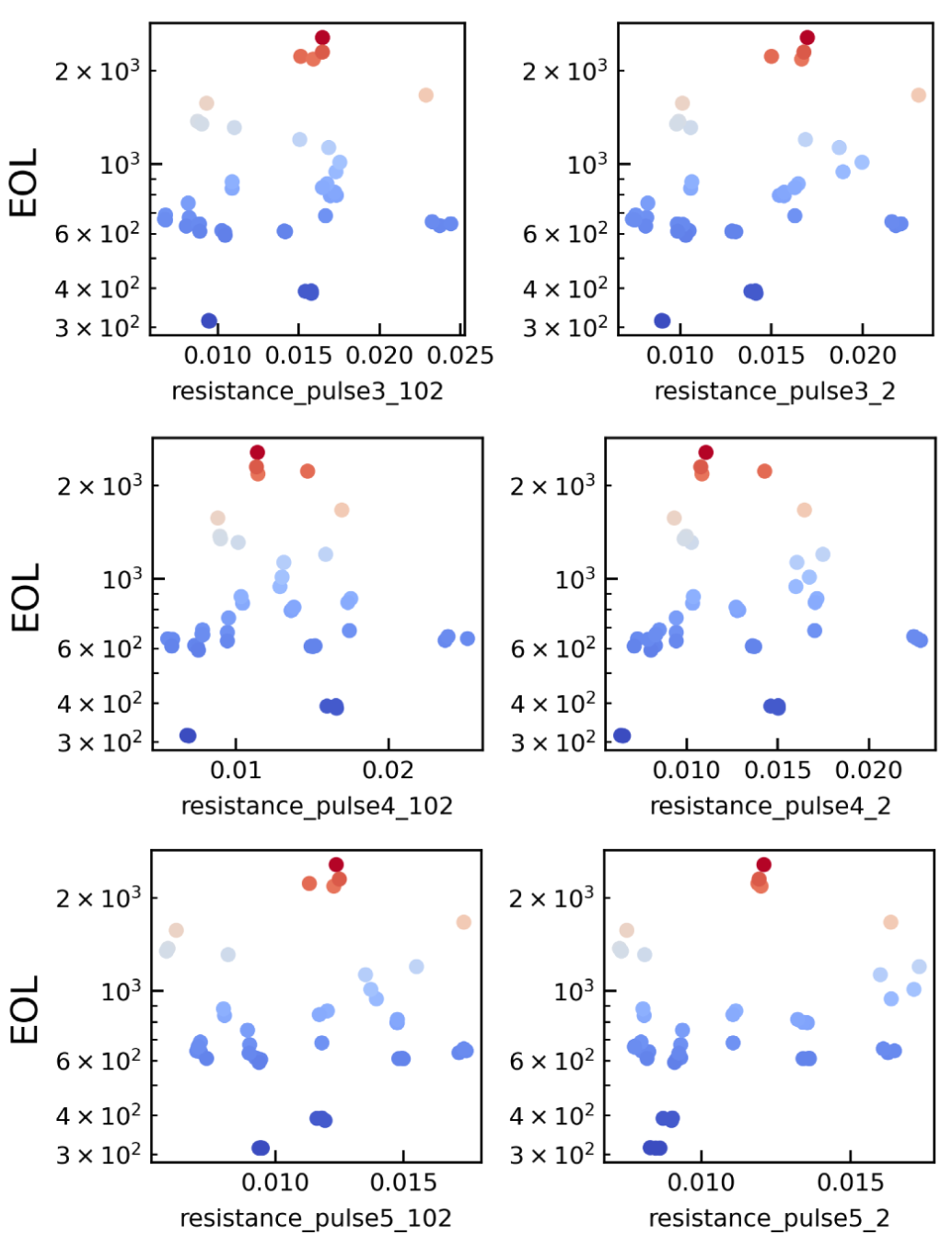}
    \caption{Feature correlation of the DCIR resistances of cycle 102 and 2 for pulse3, pulse4, and pulse5.}
    \label{fig:pairplots_res_pulse345}
\end{figure}

\begin{figure}[h]
    \centering
    \includegraphics[width=0.8\textwidth]{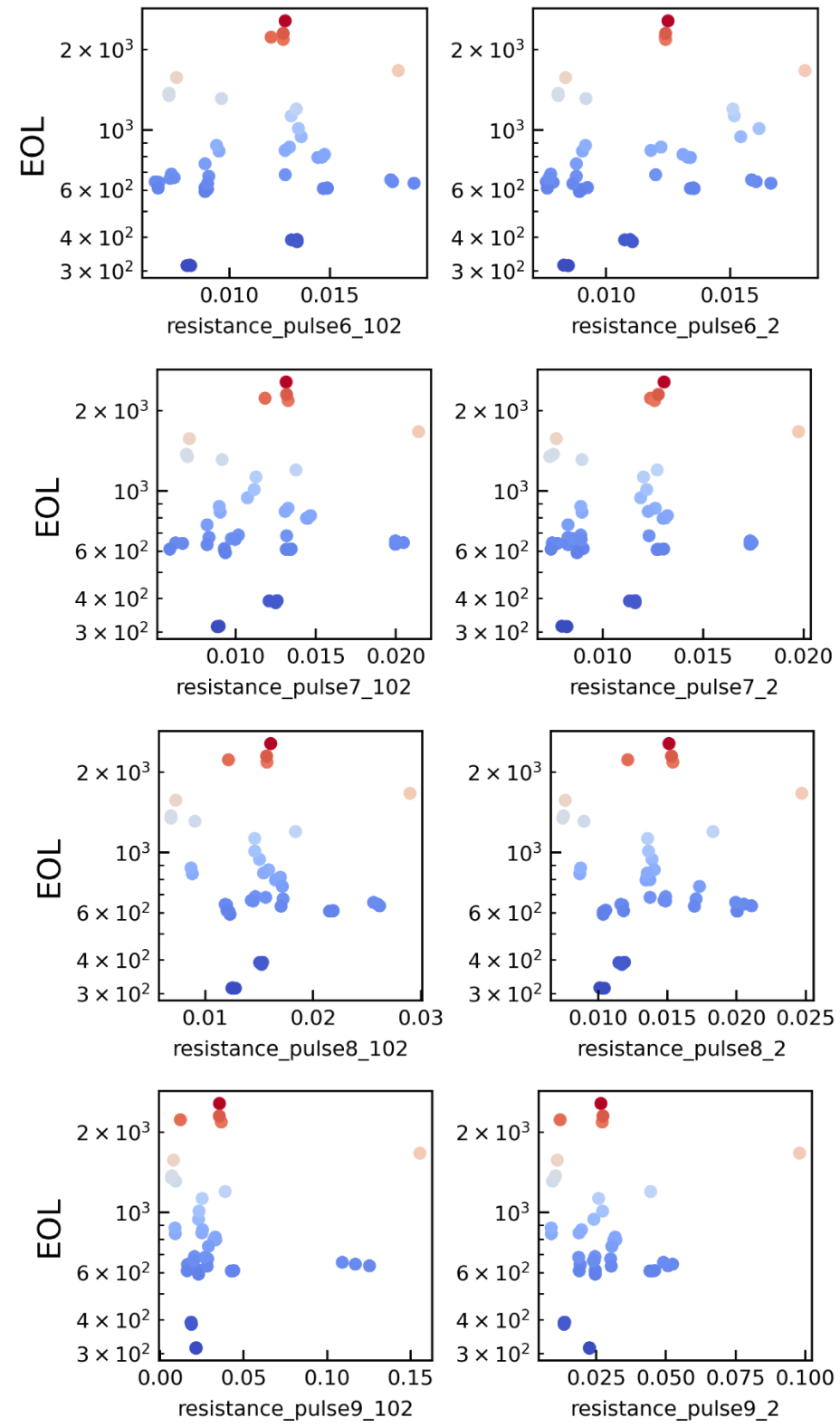}
    \caption{Feature correlation of the DCIR resistances of cycle 102 and 2 for pulse6, pulse7, pulse8, and pulse9.}
    \label{fig:pairplots_res_pulse6789}
\end{figure}

Feature correlation (on a log scale) for all features considered in this work are shown in Figs \ref{fig:pairplots_dqv} - \ref{fig:pairplots_res_pulse6789}. Generally speaking, we do not observe high feature correlation with the end of life of a battery, suggesting that principal component analysis to reduce the feature space is a key component of the machine learning workflow.

\newpage

\end{document}